\documentclass[runningheads]{llncs}

\usepackage{eccv}

\usepackage{eccvabbrv}

\usepackage{graphicx}
\usepackage{booktabs}

\usepackage[accsupp]{axessibility}  

\usepackage{hyperref}

\usepackage{orcidlink}
\usepackage{multirow}
\usepackage{makecell}
\usepackage{diagbox}
\usepackage{threeparttable}

\begin{document}

\title{MagicEraser: Erasing Any Objects via Semantics-Aware Control} 


\author{Fan Li\inst{1*}\orcidlink{0000-0002-3595-8361}
\and Zixiao Zhang\inst{1*}
\and Yi Huang\inst{2}\orcidlink{0000-0002-8443-6877} 
\and Jianzhuang Liu\inst{2}\orcidlink{0000-0002-7960-9382}
\and Renjing Pei\inst{1}
\and\\ Bin Shao\inst{1}
\and Songcen Xu\inst{1}
} 

\authorrunning{F.~Li et al.}

\institute{Huawei Noah’s Ark Lab \\
\and Shenzhen Institute of Advanced Technology, Chinese Academy of Sciences\\
\email{\{lifan61, zhangzixiao3, peirenjing, shaobin3, xusongcen\}@huawei.com, \{yi.huang, jz.liu\}@siat.ac.cn}\\
* Equal Contribution
}

\maketitle
	
\begin{abstract}
The traditional image inpainting task aims to restore corrupted regions by referencing surrounding background and foreground. However, the object erasure task, which is in increasing demand, aims to erase objects and generate harmonious background. Previous GAN-based inpainting methods struggle with intricate texture generation. Emerging diffusion model-based algorithms, such as Stable Diffusion Inpainting, exhibit the capability to generate novel content, but they often produce incongruent results at the locations of the erased objects and require high-quality text prompt inputs. To address these challenges, we introduce MagicEraser, a diffusion model-based framework tailored for the object erasure task. It consists of two phases: content initialization and controllable generation. In the latter phase, we develop two plug-and-play modules called prompt tuning and semantics-aware attention refocus. Additionally, we propose a data construction strategy that generates training data specially suitable for this task. MagicEraser achieves fine and effective control of content generation while mitigating undesired artifacts. Experimental results highlight a valuable advancement of our approach in the object erasure task.
\end{abstract}

\section{Introduction}
\label{sec:intro}
Image inpainting is a long-standing task that originally completes erased or corrupted regions within an image by incorporating information from their surrounding background and foreground. However, our focus extends beyond traditional inpainting to a more nuanced task—object erasure. While traditional inpainting aims to restore missing or damaged parts, our objective is to generate harmonious background after removing specific objects. The generative models utilized in both tasks share certain similarities, prompting us to delve into the evolution of image inpainting. Subsequently, we identify the challenges posed by existing inpainting algorithms when applied to the object erasure task. 

\begin{figure*}[t]
	\centering
	\begin{subfigure}[t]{0.2435\textwidth}
		\centering
	\includegraphics[width=3.055cm]{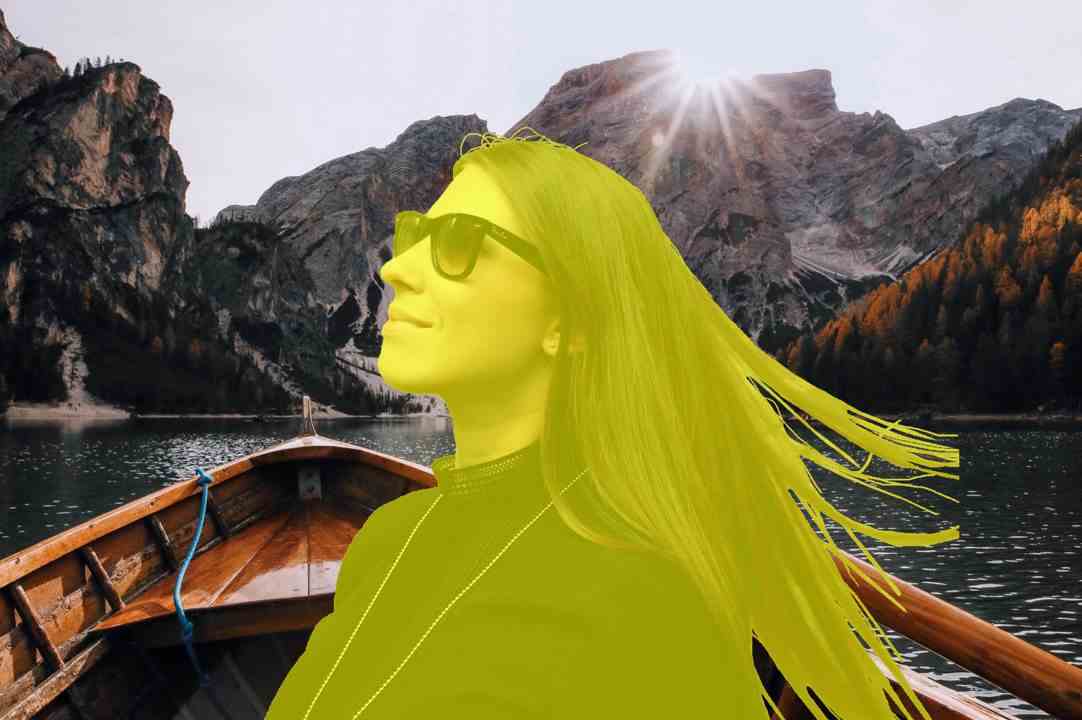}
		\caption{Input+Mask}
	\end{subfigure}
	\begin{subfigure}[t]{0.2435\textwidth}
		\centering
		\includegraphics[width=3.055cm]{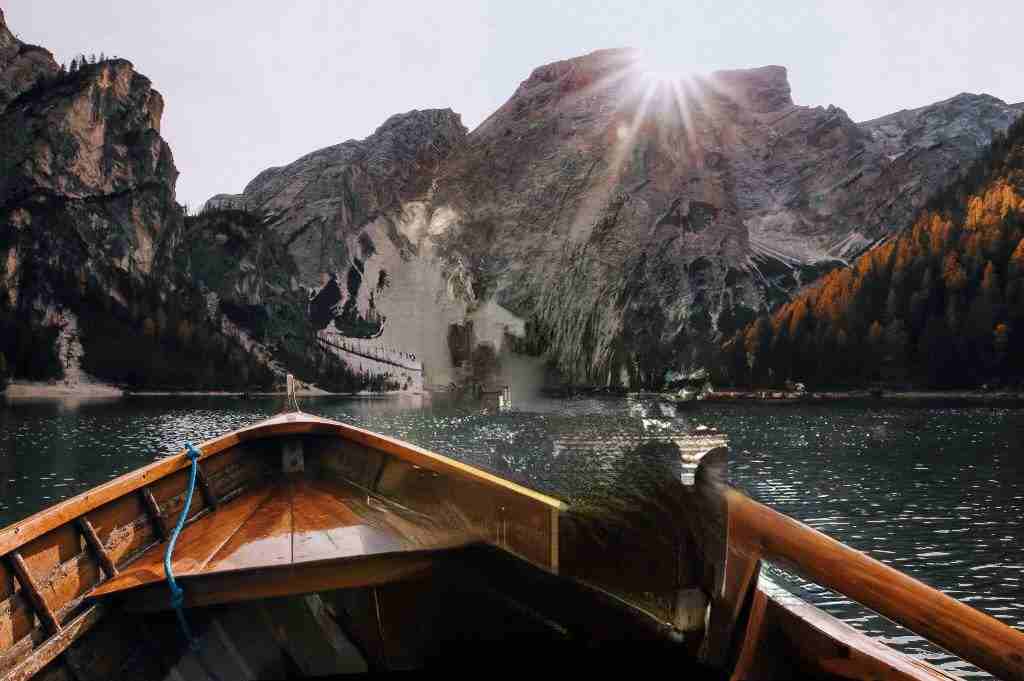}
		\caption{MAT}
	\end{subfigure}
	\begin{subfigure}[t]{0.2435\textwidth}
		\centering
		\includegraphics[width=3.055cm]{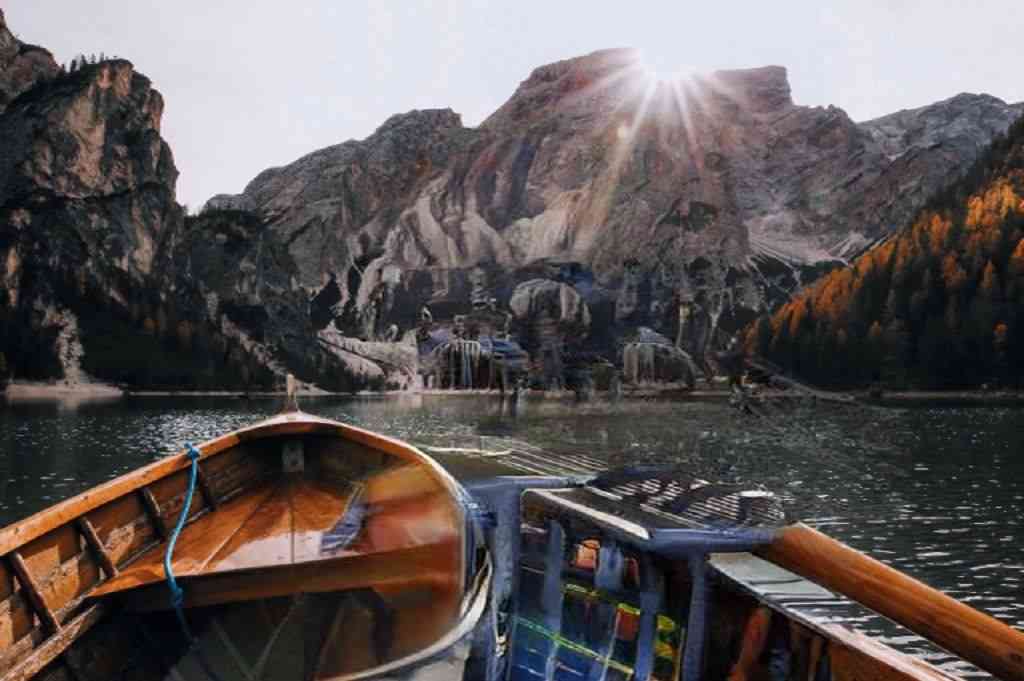}
		\caption{Co-Mod}
	\end{subfigure}
	\begin{subfigure}[t]{0.2435\textwidth}
		\centering
		\includegraphics[width=3.055cm]{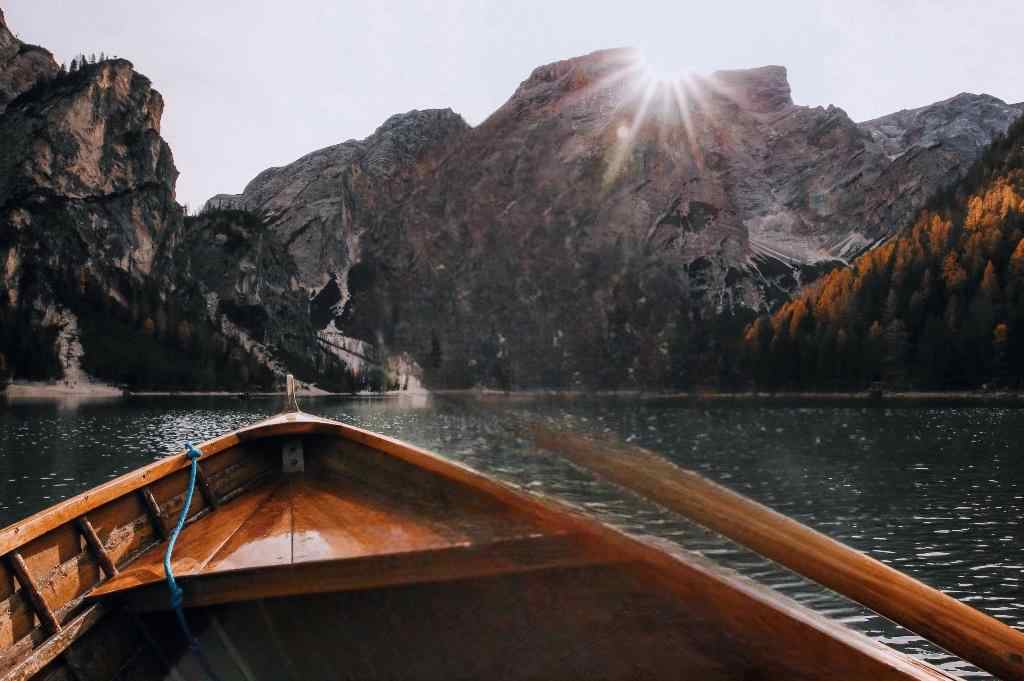}
		\caption{LaMa}
	\end{subfigure}
	\hfill
	\begin{subfigure}[t]{0.2435\textwidth}
		\centering
		\includegraphics[width=3.055cm]{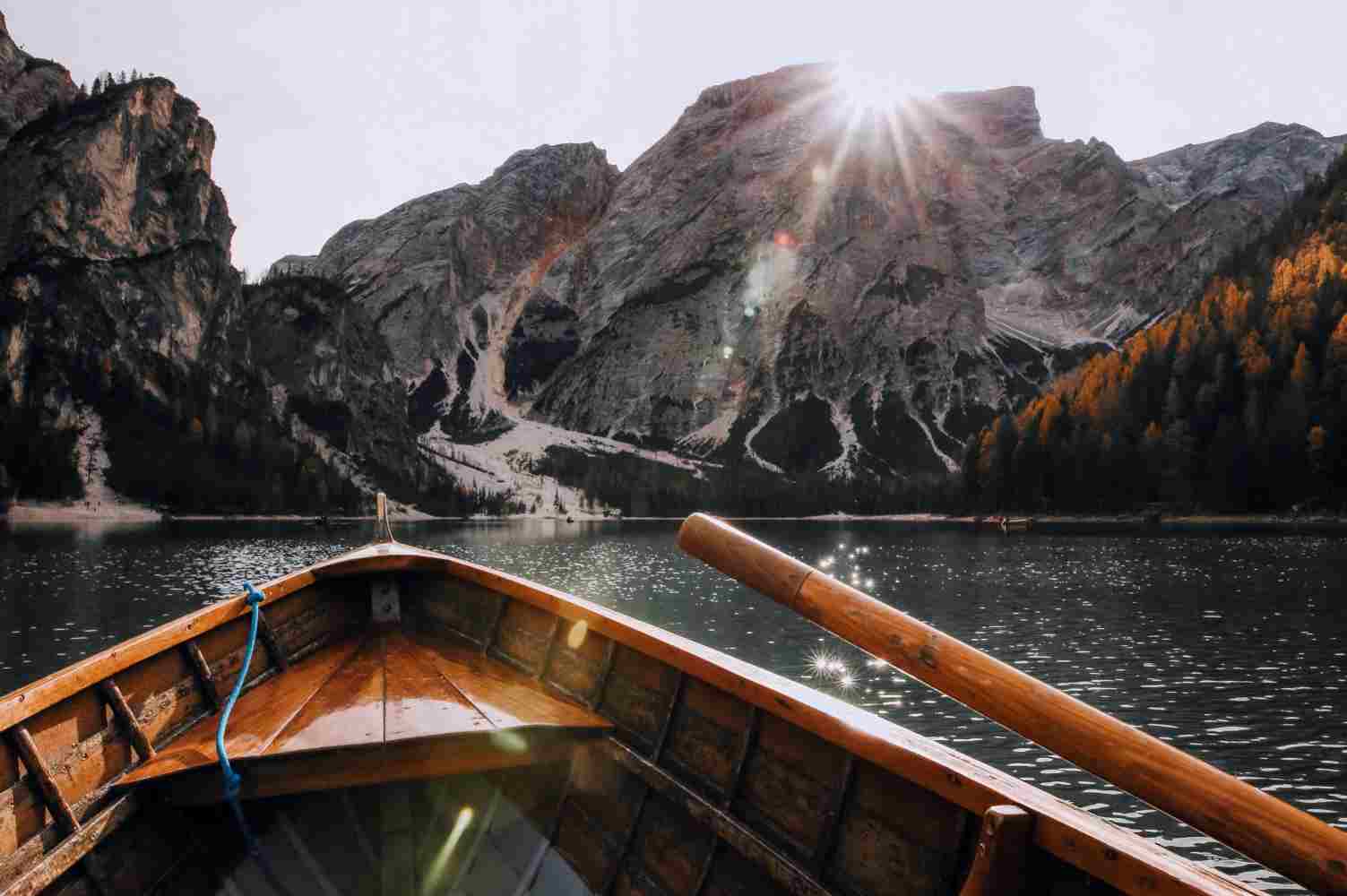}
  \caption{Reference}
	\end{subfigure}
	\begin{subfigure}[t]{0.2435\textwidth}
		\centering
		\includegraphics[width=3.055cm]{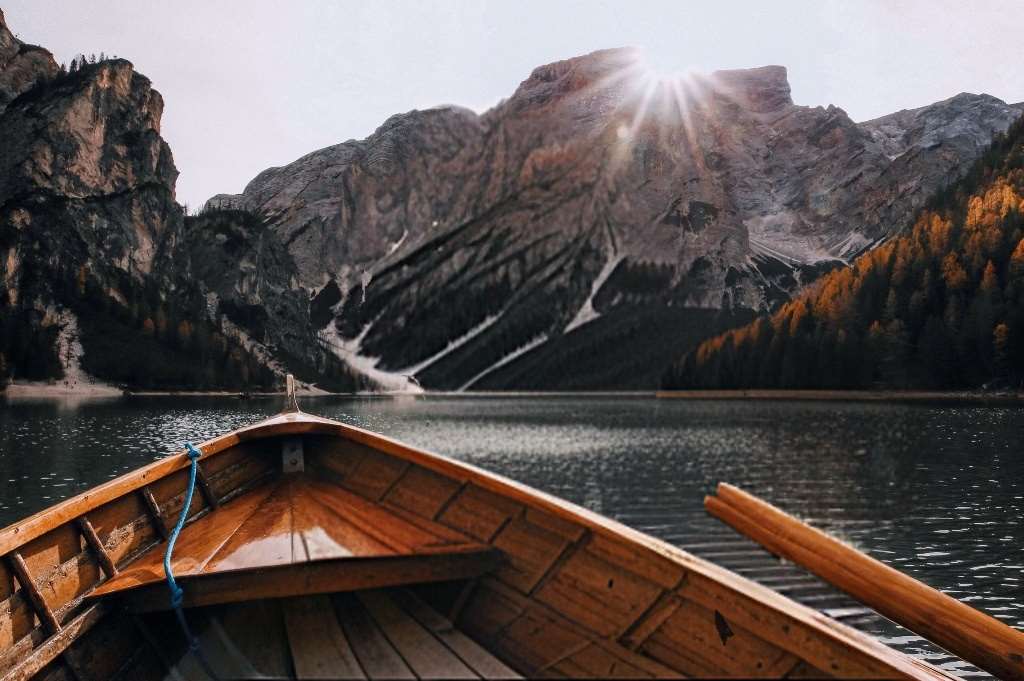}
  \caption{MagicEraser}
	\end{subfigure}
 	\begin{subfigure}[t]{0.2435\textwidth}
		\centering
		\includegraphics[width=3.055cm]{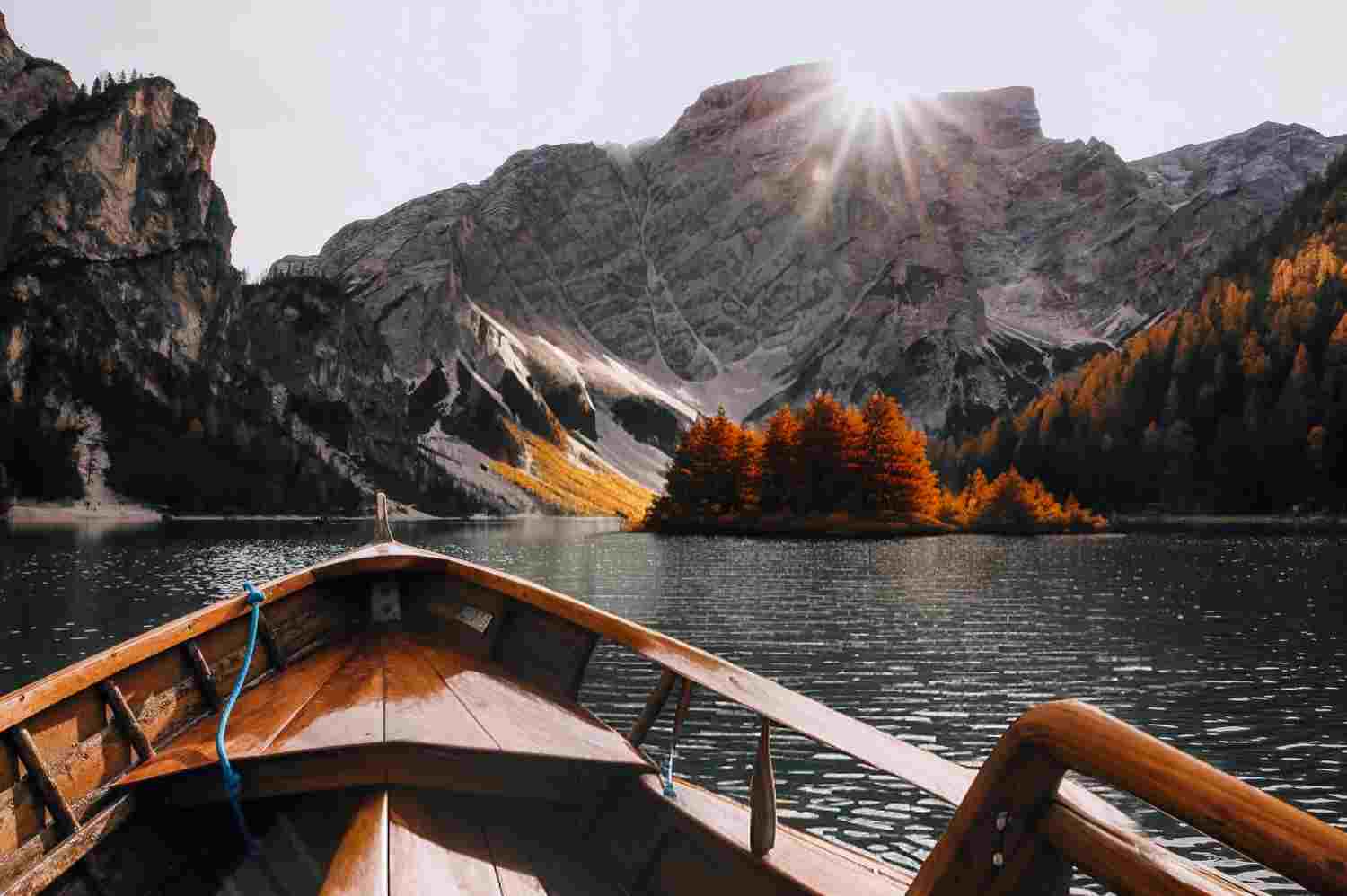}
  \caption{SD Inpainting}
	\end{subfigure}
	\begin{subfigure}[t]{0.2435\textwidth}
		\centering
		\includegraphics[width=3.055cm]{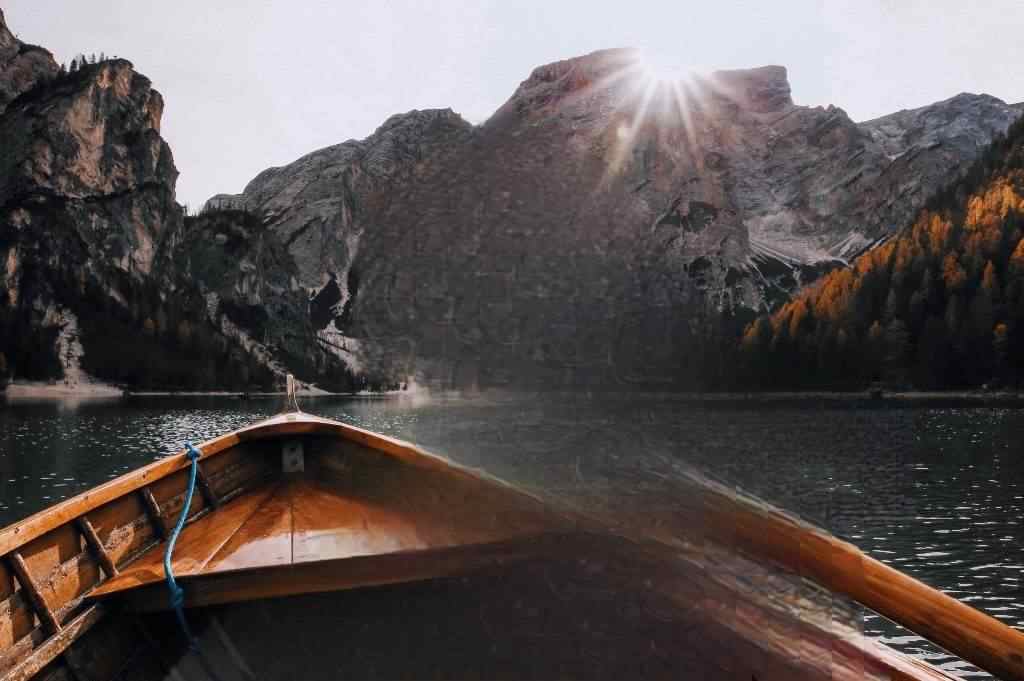}
  \caption{CoordFill}
	\end{subfigure}
	\caption{Comparison with five state-of-the-art inpainting algorithms: MAT~\cite{li2022mat}, Co-Mod~\cite{zhao2021large}, LaMa~\cite{suvorov2022resolution}, CoordFill~\cite{liu2023coordfill} and Stable Diffusion (SD) Inpainting~\cite{rombach2022high}. MagicEraser can effectively erase masked objects and achieve the best texture consistency and content fidelity.}
	\label{fig:teaser}
\end{figure*}

Earlier methods in image inpainting, particularly those relying on generative adversarial networks (GANs)~\cite{goodfellow2014generative, li2023image}, encounter challenges in generating high-quality textures for large corrupted regions and struggle with object erasure, which includes approaches like LaMa~\cite{suvorov2022resolution}, which introduces large mask inpainting based on fast Fourier convolutions (FFCs), and CoordFill~\cite{liu2023coordfill}, which utilizes parameterized coordinate querying and convolution simplification tricks for efficient high-resolution image inpainting. Both MAT~\cite{li2022mat} and Co-Mod~\cite{zhao2021large} aim to enhance the performance of inpainting large regions of missing information in images. Despite their advancements, these GAN-based methods still encounter difficulties in generating high-quality textures for complex backgrounds, particularly in large erased regions, as shown in the example of Fig.~\ref{fig:teaser}.

Recently, diffusion models~\cite{ho2020denoising, lugmayr2022repaint}, including Stable Diffusion~\cite{rombach2022high}, DALL-E~\cite{ramesh2022hierarchical, ramesh2021zero}, and Imagen~\cite{ho2022imagen}, have shown promise in text-to-image generation. Meanwhile, approaches like GLIDE~\cite{nichol2022glide} and Stable Diffusion Inpainting~\cite{rombach2022high} (inpainting version of Stable Diffusion) fine-tune diffusion models with random masks, recovering missing regions conditioned on corresponding image captions. Particularly, when applied to inpainting, diffusion models substitute random noise in the background with a noisy version of the original image during the reverse diffusion process. However, this method often yields random and undesirable outcomes due to its heavy reliance on high-quality text prompts. For the example in Fig.~\ref{fig:teaser}, SD Inpainting generates a relatively harmonious result based on this long prompt: ``\textit{The boat is on a serene lake surrounded by dramatic mountains with rugged textures. The sun is shining directly above the mountain peaks, creating a flare effect in the camera lens. There's a reflection of the sun on the water, suggesting it's a clear day. The trees on the mountainside are tinged with autumn colors, which adds warmth to the scene}''. If we use a short prompt like ``\textit{A boat on the lake}'', SD Inpainting tends to generate another different boat. Hence, controlling the generation of masked regions becomes a major challenge.

A high-quality image caption is essential for effective control and tends to emphasize global image features, but it is not very user-friendly. There may be semantic misalignment between the locally erased content and the global text description, potentially leading diffusion models to fill the masked regions with object-level foreground rather than the surrounding background, as illustrated by Stable Diffusion Inpainting in Fig.~\ref{fig:teaser}. This emphasizes the necessity for a more nuanced and user-friendly approach to the object erasure task.

To address the aforementioned challenges, we introduce MagicEraser, a new user-friendly diffusion model-based framework designed for object erasure. Broadly, we break down the process into two phases: content initialization and controllable generation. In the former, we employ a pretrained traditional inpainting method to initialize the content within masked regions. The latter has two plug-and-play modules named prompt tuning and semantics-aware attention refocus. The prompt tuning module, employing textual inversion\cite{gal2022textual} and LoRA\cite{hu2021lora} fine-tuning techniques, primarily aims to preserve the capability of multi-modal understanding without requiring manual input prompts. This dramatically improves the usage for ordinary people in practical applications. 
On the other hand, the semantics-aware attention refocus module is effective and training-free. It utilizes semantic cues obtained through panoptic segmentation and then adaptively adjusts the attention values of the background and foreground. This adaptive adjustment contributes to enhanced controllability of the generation process. Additionally, different from traditional training data construction for inpainting, we propose a new data construction strategy for fine-tuning the diffusion model. Experimental results highlight a valuable advancement of our approach in the object erasure task across various scenarios.

Our contributions are summarized in the following: (1) We propose MagicEraser, an effective and user-friendly object-erasing framework based on the diffusion model. (2) We introduce a data construction strategy specifically designed for the object erasure task. (3) We present prompt tuning and training-free semantics-aware attention refocus to enhance the controllability of the generation process. (4) Comprehensive experiments validate that MagicEraser achieves state-of-the-art quantitative and qualitative results.

\section{Related Work}
\subsection{GAN-Based Image Inpainting}
Object erasure refers to removing objects from an image and restoring the background behind them, and is often considered a context-driven type of image inpainting. 
Earlier methods for this task predominantly rely on Generative Adversarial Networks (GANs) \cite{goodfellow2014generative, suvorov2022resolution} that are trained on massive datasets. 
For instance,  Co-Mod~\cite{zhao2021large} harnesses the generative capability of unconditional modulation techniques~\cite{karras2019stylebased, karras2020analyzing} and employs co-modulation of both conditional and stochastic style representations to handle large-scale missing regions. 
LaMa~\cite{suvorov2022resolution} utilizes the Fast Fourier Convolutions (FFCs) to extend the network's receptive field across the entire image at early stages, thereby enhancing perceptual quality and facilitating adaptation to high-resolution images that are not seen during training.
MAT~\cite{li2022mat} presents a transformer-based framework designed for high-resolution inpainting. However, its practical application is limited by the inefficient multi-stage structure.
CoordFill~\cite{liu2023coordfill}  proposes a more efficient inpainting decoder utilizing an implicit representation with a multi-layer perceptron (MLP) network.
These methods primarily excel in scenarios involving simple backgrounds with repetitive textures, such as grass or sky. However, complex backgrounds characterized by inconsistent textures or lighting conditions pose significant challenges, often leading them to generating content that lacks consistency and exhibits noticeable blurriness. This paper tackles the problem based on diffusion models with semantic awareness of context.

\subsection{Diffusion Model-Based Image Inpainting}
Recent years have seen a growing interest in diffusion models \cite{sohl2015deep, ho2020denoising, song2021score, song2021denoising} across various vision tasks such as image generation \cite{ho2022cascaded, rombach2022high, saharia2022photorealistic, nichol2022glide, cao2024survey, croitoru2023diffusion}, editing \cite{meng2021sdedit, huang2024diffusion, hertz2022prompt, brooks2023instructpix2pix, wang2023dynamic} and restoration \cite{ozdenizci2023restoring, kawar2022denoising, huang2024wavedm, li2023diffusion} due to their superior capacity to capture complex data distributions and more stable training than GANs. Similar to GAN-based approaches, early studies utilizing diffusion models for inpainting primarily focus on leveraging the surrounding context to fill the missing pixels. For instance, Palette \cite{saharia2022palette} trains a diffusion model by directly concatenating masked images with their original versions as input. Repaint \cite{lugmayr2022repaint} blends masked regions generated from a pretrained unconditional diffusion model and unmasked regions from the original images at each sampling step.  However, these methods often fall short in offering precise control over the generated content.

With the advance of text-to-image (T2I) diffusion models, this limitation is being mitigated by incorporating additional conditions, such as text, segmentation maps, and reference images. For instance, Stable Diffusion Inpainting, an adaptation from Stable Diffusion \cite{rombach2022high}, finetunes the pretrained T2I model using randomly generated masks, masked images, and the captions of original images. This approach, however, sometimes fails to maintain relevance to the text prompts, particularly with small masked regions or when only part of an object is covered. 
To enhance the precision of inpainted content, SmartBrush \cite{xie2023smartbrush} introduces a precision factor, enabling the generation of masks ranging from fine to coarse by applying Gaussian blur to accurate instance masks. Imagen Editor \cite{wang2023imagen} extends the Imagen \cite{saharia2022photorealistic} model through finetuning with precise object masks, dynamically generated by an object detector, SSD Mobilenet v2 \cite{sandler2018mobilenetv2}, rather than random masking. Although these advancements improve the fidelity of generated content, they tend to introduce new objects into the masked regions rather than restoring the original background, which is crucial for the object erasure task. 

Addressing this challenge, specific approaches have been tailored for precise object erasure.
Inst-Inpaint \cite{yildirim2023inst} allows removal of objects specified by text instructions, bypassing the need for binary masks. It trains a diffusion model on the self-constructed GQA dataset comprising source images, their ground truths with objects removed, and text instructions. MagicRemover \cite{yang2023magicremover} employs an attention guidance strategy within the diffusion model's sampling process to facilitate the erasure of inpainting regions and the restoration of occluded content. PowerPaint \cite{zhuang2023task} finetunes a T2I model with dual task prompts, $P_{obj}$ for text-guided object inpainting and $P_{ctxt}$ for context-aware image inpainting, where $P_{obj}$ serves as a negative prompt with classifier-free guidance sampling for object removal.
Despite these advancements, challenges still remain when they confront complex backgrounds, often resulting in unnaturally generated content. This study seeks to overcome such obstacles through semantics-aware control and the construction of high-quality training data.

\section{Preliminaries}
\subsection{Diffusion Models}
Denoising Diffusion Probabilistic Models (DDPMs) \cite{ho2020denoising} define a forward noising process following the Markov chain that transforms a data sample $x_0$ from its real data distribution $q(x)$ into a sequence of noisy samples $x_t$ in $T$ steps with a variance schedule $\beta_1,\ldots,\beta_T$: $q(x_{t}\vert x_{t-1}) = \mathcal{N}(x_{t};\sqrt{1-\beta_t}x_{t-1},\beta_t \mathbf{I})$. 
The closed form of the forward process can be expressed as $x_t = \sqrt{\bar{\alpha}_t}x_0 + \sqrt{1-\bar{\alpha}_t}\epsilon,$
where $\alpha_t=1-\beta_t$, $\bar{\alpha}_t=\prod_{i=1}^t \alpha_i$, and $\epsilon \sim \mathcal{N}(\mathbf{0}, \mathbf{I})$.

To generate images starting with a noisy sample from the standard Gaussian distribution $\mathcal{N}(\mathbf{0}, \mathbf{I})$, diffusion models learn to reverse the above process through a joint distribution $p_{\theta}(x_{0:T})$ that follows the Markov chain with parameters $\theta$: $p_{\theta}(x_{t-1}\vert x_t) = \mathcal{N}(x_{t-1};\mu_{\theta}(x_t,t), \Sigma_{\theta}(x_t,t)).$
The parameters $\theta$ are usually optimized by a neural network $\epsilon_{\theta}(x_t,t)$ that directly predicts noise vectors $\epsilon_t$ instead of $\mu_{\theta}$ and $\Sigma_{\theta}$ with the following simplified objective \cite{ho2020denoising}:
\begin{equation}
	L_{simple} =\mathbb{E}_{x_0,t,\epsilon_t\sim\mathcal{N}(\mathbf{0},\mathbf{I})}\Big[\vert\vert\epsilon_t - \epsilon_{\theta}(x_t,t)\vert\vert^2 \Big].
	\label{eq:training_obj}
\end{equation}



As for conditional diffusion models, \eg, T2I generation and inpainting models, the conditions, \eg, text and mask, can be fed into the network $\epsilon_{\theta}$ without changing the loss function. Then the model learns to generate images that are consistent with the conditions.

\subsection{Stable Diffusion Inpainting}
Stable Diffusion Inpainting, a variant of Stable Diffusion \cite{rombach2022high},  is specifically finetuned for the image inpainting task using a randomly generated mask, the corresponding masked image, and the caption of the complete image. This adaptation enables the model to utilize information from the unmasked regions effectively during its training phase. Unlike the original Stable Diffusion which processes a 4-channel noisy latent $ z_{t} \in \mathbb{R}^{h\times w\times 4}$ in the Variational Autoencoder (VAE) \cite{kingma2014auto} latent space, Stable Diffusion Inpainting adapts the first convolutional layer of the denoising network to accept a 9-channel input. This expanded input $ z'_{t} \in \mathbb{R}^{h\times w\times 9} $ is the concatenation of the masked image latent $z_{masked} \in \mathbb{R}^{h\times w\times 4} $, $ z_{t} \in \mathbb{R}^{h\times w\times 4}$, and the corresponding randomly generated mask $m \in \mathbb{R}^{h\times w\times 1}$. Therefore, the optimization loss of the 9-channel Stable Diffusion is: 
\begin{equation}
	L_{9ch} =\mathbb{E}_{z_0, z_{masked}, m,t,y,\epsilon_t\sim\mathcal{N}(\mathbf{0},\mathbf{I})}\Big[\vert\vert\epsilon_t - \epsilon_{\theta}(z'_t,t,\tau(y),m)\vert\vert^2 \Big],
	\label{eq:loss_sd_inpt}
\end{equation}
where $\tau(\cdot)$ is a text encoder that maps a text prompt $y$ into a conditional vector.

\section{Approach}
Given an image $x$ and a binary mask $m$ indicating the target objects for erasure, our objective is to generate an image $\hat{x}$ where the masked regions are seamlessly replaced with harmonious background without introducing new foreground objects. Furthermore, we aim for  this erasing process to be achieved without the necessity for extra manual text prompt input.

\begin{figure*}[t]
	\centering 
	\includegraphics[width=1\linewidth]{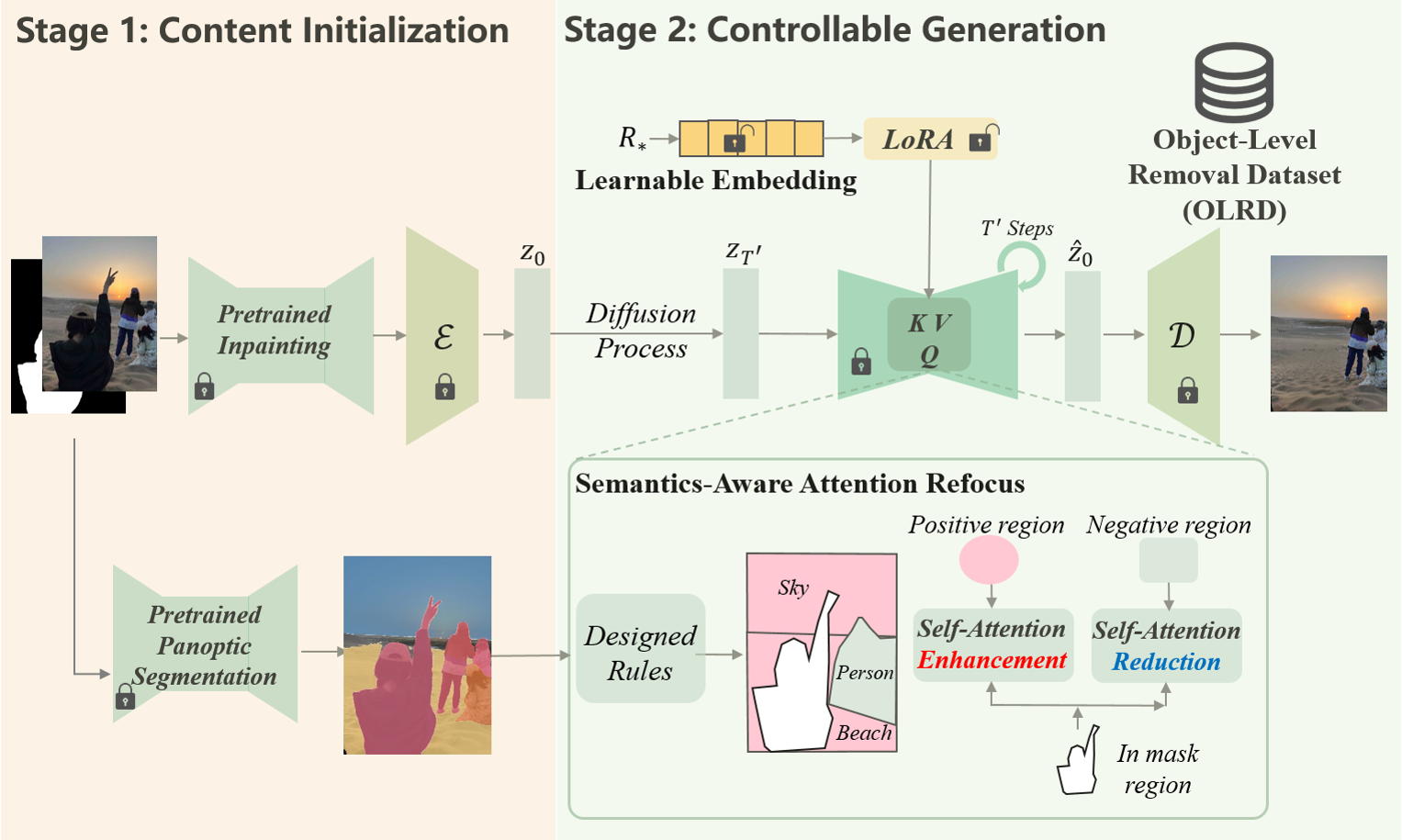}
	\caption{MagicEraser, built upon Stable Diffusion Inpainting, comprises two main stages: content initialization and controllable generation. Additionally, we construct an object-level removal dataset (OLRD) specifically designed for the object erasure task.
	}
	\label{framework}
\end{figure*}

\label{sec:blind}
\subsection{Overall Framework}
Existing diffusion model-based inpainting models, similar to their text-to-image counterparts, often heavily rely on high-quality text prompt input~\cite{rombach2022high, meng2021sdedit}, which is not intuitive to obtain, especially for ordinary users. These models demonstrate limitations in understanding multi-modal inputs, struggling to interpret semantic information and to achieve seamless background completion. Therefore, we propose a diffusion model-based framework, MagicEraser, as highly suitable and user-friendly for the object erasure task. Illustrated in Fig.~\ref{framework}, MagicEraser comprises two main phases: content initialization and controllable generation. The former initializes the content of the erasure regions, while the latter governs denoising generation based on learnable text prompts and training-free semantics-aware attention refocus. Additionally, we propose a specialized data construction strategy for fine-tuning the diffusion model. 

\subsection{Content Initialization}\label{sec:init}
Latent initialization plays a crucial role in high-resolution image generation, particularly within the latent diffusion model (LDM)~\cite{rombach2022high}. The noising and denoising processes typically take places in the latent space $\mathcal{Z}$, with a parameter called denoising strength ($s\in(0,1]$) controlling the entire procedure. Specifically, in the sampling process of the diffusion model with a maximum of $T$ sampling steps, the actual number of sampling steps is given by $T'=\lfloor T \cdot s \rfloor$. This means that the denoising process starts from $z_{T'}$, which can be calculated as: 

\begin{equation}
	z_{T'} = \sqrt{\bar{\alpha}_{T'}}z_0 + \sqrt{1-\bar{\alpha}_{T'}}\epsilon,
	\label{eq:latent_noising}
\end{equation}
where $z_0=E(x_0)$,  $x_0$ is the given image and $E$ is the encoder of Variational Autoencoder (VAE)~\cite{kingma2014auto} within LDM.

When $s=1$, the generation starts from standard Gaussian noise $z_T\sim\mathcal{N}(\mathbf{0},\mathbf{I})$, often resulting in significant deviations from the original image $x_0$. In the object erasure task, where the objective is to generate a harmonious background, a smaller $s$ (e.g., $s=0.75$) is capable of enhancing texture harmony. However, this can lead to the generation of undesired new objects similar to those being erased in the masked regions.

To address this problem, we employ a pretrained traditional inpainting model such as LaMa or CoordFill to roughly initialize the content of the erasing regions in the pixel space. The pre-processed image $\tilde x_0$ is then passed through the VAE encoder to obtain $\tilde z_0=E(\tilde x_0)$. Subsequently, the initial noisy latent vector $z_{T'}$ can be calculated by Eq.~\ref{eq:latent_noising}. Under this initialization, we set $s=0.9$ in MagicEraser, maintaining fine and harmonious texture while mitigating the generation of undesired artifacts.

\subsection{Controllable Generation}\label{sec:generation}
\subsubsection{Prompt Tuning.}
\label{Prompt_Tuning}
The growing need for object removal in photography is primarily driven by ordinary people. They often lack the expertise to acquire professional-grade prompts (see the footnote on page 2), which are essential for accurately directing diffusion models to remove unwanted objects. To address this, we design a prompt tuning method for object erasure based on our object-level removal dataset (OLRD\footnote{https://github.com/lifan724/magic\_eraser}) detailed in Section~\ref{sec:data}, which only tunes a small amount of the parameters added to the U-Net in Stable Diffusion Inpainting, avoiding destroying the capability of the pre-trained model.

Specifically, our objective is to obtain a tuned prompt, which can teach the model a new concept, ``background completion'', using Textual Inversion~\cite{gal2022textual}. We designate a placeholder string ``$R_*$'' to represent this new concept, whose corresponding token embedding added to the vocabulary is denoted as $v_*$. We initialize $v_*$ using Textutal Inversion on a small random subset of OLRD. To condition the generation, we utilize the background tag (e.g., ``sky'' or ``beach'') to obtain a short text prompt $y$ in the form of ``A photo of $R_*$ sky'' or ``A photo of $R_*$ beach'', where the tag is from the results of a pretrained panoptic segmentation algorithm. Note that $R_*$ can be considered as a \textit{universal} ``background completion'' concept that is expected to force the model to focus more on background generation (e.g., ``sky'' or ``beach'', etc.). If we only use the text prompt without $R_*$ (e.g., ``A photo of sky'', ``A photo of beach'', etc.), the diffusion model tends to generate new objects similar to those to be erased.

We find that relying solely on Textual Inversion is not enough to capture this intricate concept. So we further tune it together with the model fine-tuning on OLRD using the low-rank method LoRA~\cite{hu2021lora}. The additional parameters $\phi$ of LoRA are added to the U-Net of the diffusion model and simultaneously trained with $v_*$, enhancing the model's understanding of the concept of object erasure. The optimization of the diffusion model fine-tuning is then defined as: 
\begin{equation}
	v_*, \phi_* = \arg\min_{v, \phi} \mathbb{E}_{z_0, z_{masked}, m,t,y,\epsilon_t\sim\mathcal{N}(\mathbf{0},\mathbf{I})}\Big[\vert\vert\epsilon_t - \epsilon_{\theta, \phi}(z'_t, t, \tau(y),m)\vert\vert^2 \Big].
	\label{eq:loss_v_lora}
\end{equation}
Additionally, to avoid degrading the generation quality by only using the above simple text prompt during fine-tuning (e.g., ``A photo of $R_*$ sky''). We only use them with a $50\%$ chance and use image captions detailed in Section~\ref{sec:data} with another $50\%$ chance. During inference, the model only uses the above simple text prompts that are automatically constructed by the panopatic segmentation algorithm and the learned $R_*$ ($v_*$), without the need of user input.

\subsubsection{Semantics-Aware Attention Refocus.}\label{sec:sem_attn}The self-attention layers in Stable Diffusion are crucial components that reorganize intermediate features to ensure globally coherent generated content. Previous research~\cite{yang2023magicremover, kim2023dense, epstein2023diffusion} has demonstrated that appropriately modulating the self-attention layers can enhance the controllability of T2I models. In the context of object erasure, pixels outside the mask can be considered as a type of "visual prompts", influencing the content generation within the mask. Therefore, modulating self-attention layers to focus more on desired regions outside the mask and to ignore undesired ones can improve the generation of coherent content and suppress the generation of incongruent content. Consequently, we propose a training-free semantics-aware attention refocus module, which utilizes semantic cues obtained through panoptic segmentation as guidance for modulating the self-attention layers. Our experiments show that this module significantly enhances the controllability of our diffusion model, thereby boosting the quality of generated images.

\begin{figure}[tb]
	\centering 
\includegraphics[width=1\linewidth]{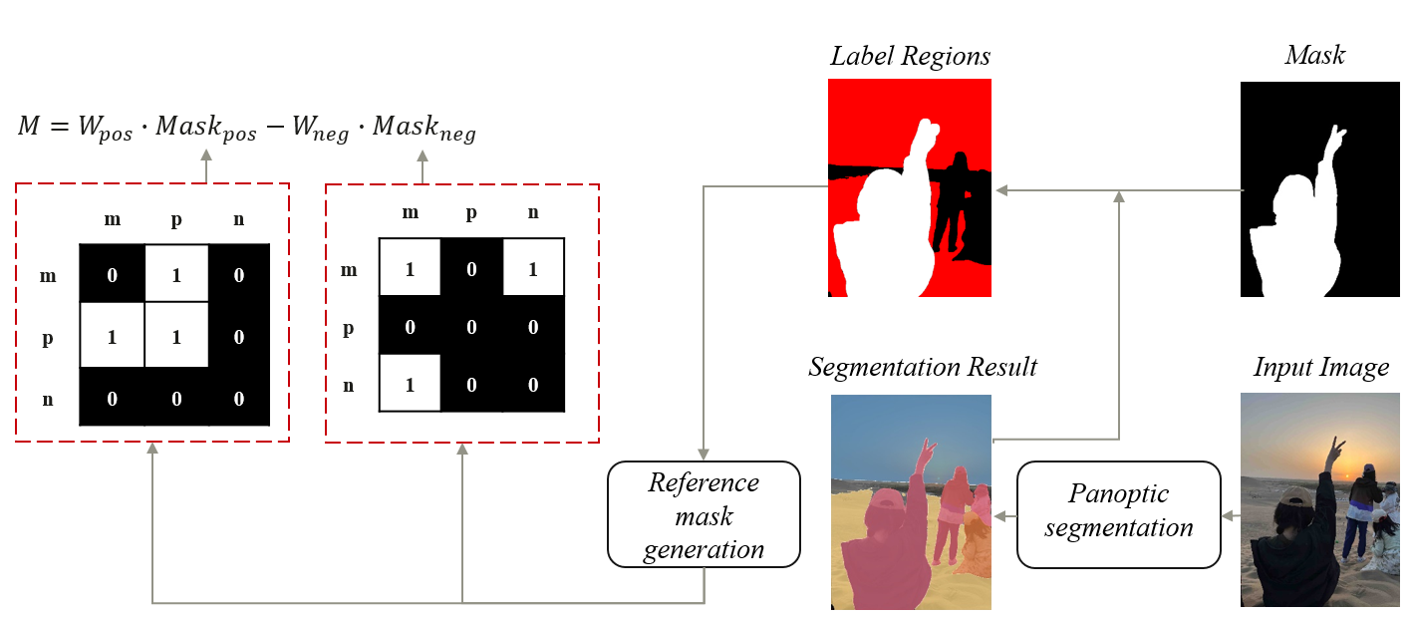}
	\caption{Semantics-aware attention refocus. We combine the panoptic segmentation result of the input image with the input mask to generate $Mask_{pos}$ and $Mask_{neg}$. With the input mask and the panoptic segmentation results, we obtain the labels ($l$) of different regions (white for \textit{mask} (\textit{m}) regions, red for \textit{positive} (\textit{p}) regions and black for \textit{negative} (\textit{n}) regions).  
	}
	\label{fig:attn_refocus}
\end{figure}

Unlike~\cite{yang2023magicremover, epstein2023diffusion} which use well-designed losses to optimize attention maps, we opt for a direct way to modify them. Inspired by~\cite{kim2023dense}, which modulates the attention values by:
\begin{equation}
\small
	A^{'} ={\rm softmax}(\frac{QK^T + M}{\sqrt{d}}),
    \label{eq:attn_modification}
\end{equation}
we design \textit{M} as follows: 
\begin{equation}
\small
	M = W_{pos}\cdot Mask_{pos} - W_{neg}\cdot Mask_{neg},
\end{equation}
where the binary masks $Mask_{pos}$, $Mask_{neg}$ $\in$ $\mathbb{R}^{|{\rm queries}|\times|{\rm keys}|}$, indicating which self-attention values should be modulated, and $W_{pos}$, $W_{neg}$ $\in$ $\mathbb{R}^{|{\rm queries}|\times|{\rm keys}|}$ are their corresponding modulation weights. 

We design $Mask_{pos}$ and $Mask_{neg}$ to be semantics-aware by utilizing the panoptic segmentation results, as shown in Fig.~\ref{fig:attn_refocus}. Specifically, based on their semantic categories and the objects to be erased, we assign each latent pixel with a label $l \in \{\textit{mask}, \textit{positive}, \textit{negative}\}$ ($\textit{m}, \textit{p}, \textit{n}$ for short) standing for \textit{mask} regions, \textit{positive} regions and \textit{negative} regions, respectively. Here, a positive region is one whose semantics belong to background, while a negative region is one whose semantics are similar to the objects to be erased. During denoising process, using Eq.~\ref{eq:attn_modification}, we increase the self-attention values of the \textit{mask} regions with the \textit{positive} regions while decreasing them with both the \textit{negative} regions and the \textit{mask} regions. Let ${l}[i]$ be the label of pixel $i$. Then for each query pixel $i$ and key pixel $j$ in the self-attention maps, we define
\begin{equation}
\small
Mask_{pos}[i,j] = 
\left\{
\begin{aligned}
	1,\quad& {\rm if}\ ({l}[i]=\textit{m}\ {\rm and}\ {l}[j]=\textit{p})\ {\rm or}\ ({l}[i]=\textit{p}\ {\rm and}\ {l}[j] \in\{\textit{m}, \textit{p}\}),\\
	0,\quad& {\rm otherwise},
\end{aligned}
\right.
\end{equation}

\begin{equation}
\small
	Mask_{neg}[i,j] = 
	\left\{
	\begin{aligned}
		1,\quad& {\rm if}\ ({l}[i]=\textit{m}\ {\rm and}\ {l}[j]\in\{\textit{m}, \textit{n}\})\ {\rm or}\ ({l}[i]=\textit{n}\ {\rm and}\ {l}[j]=\textit{m}),\\
		0,\quad& {\rm otherwise}.
	\end{aligned}
	\right.
\end{equation}
As for $W_{pos}$ and $W_{neg}$, we first apply max and min operations to the similarity matrix ${QK^T}$, obtaining the maximum and minimum values for each query, then replicate these values along the key-axis to obtain final $S_{max}$, $S_{min}$ $\in$ $\mathbb{R}^{|{\rm queries}|\times {\rm |keys|}}$, and finally define
\begin{equation}
\small
    \begin{aligned}
    	W_{pos} = (1 - \lambda_{pos})\cdot S_{min} + \lambda_{pos}\cdot S_{max},\quad
        W_{neg} = \lambda_{neg}\cdot S_{max},
    \end{aligned}
\end{equation}
where $\lambda_{pos}$ and $\lambda_{neg}$ are empirically set to 0.8 and 1.0, respectively. In addition, as discussed in~\cite{kim2023dense}, we only modulate the self-attention layers at the initial denoising steps (\textit{t} = 1 $\sim$ 0.7). 

\subsection{Training Data Construction and Model Finetuning} \label{sec:data}
\begin{figure}[t]
	\centering 
\includegraphics[width=\textwidth,height=\textheight,keepaspectratio]{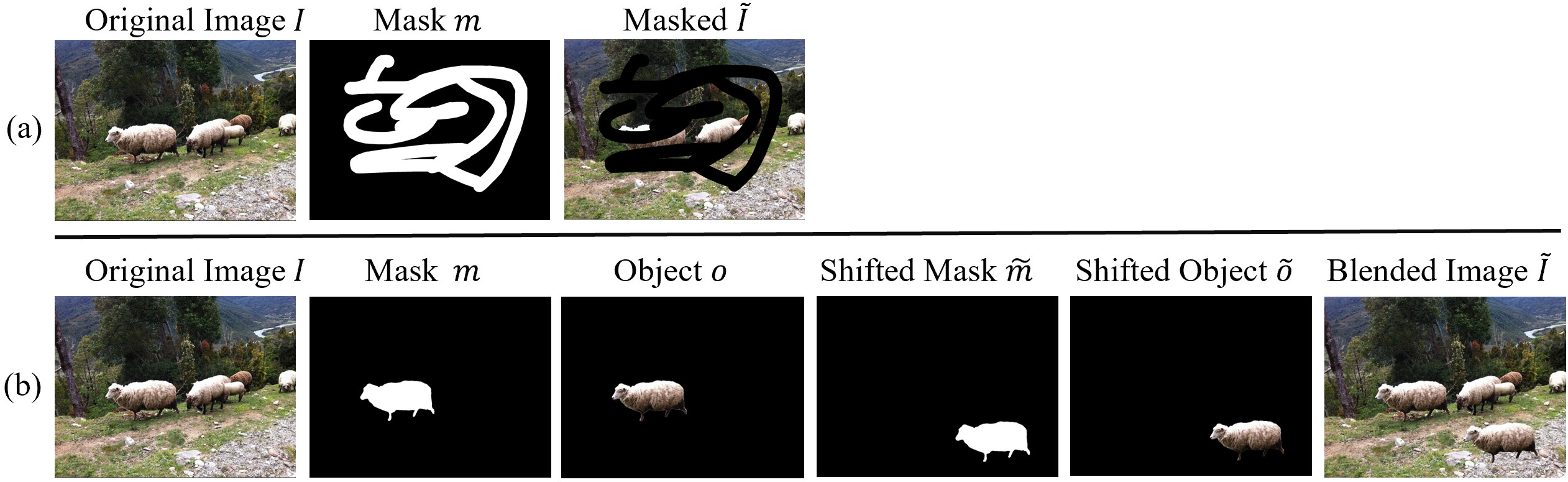}
	\caption{
		Training data comparison between the traditional inpainting and object erasure. (a) Traditional inpainting methods use random mask $m$ and the masked image $\tilde I$ to recover the original image $I$. (b) Our model uses the shifted mask $\tilde m$ and the blended image $\tilde I$ to recover $I$.
	}
	\label{fig:data}
\end{figure}

As shown in Fig.~\ref{fig:data}(a), traditional inpainting methods often generate random mask $m$ and recover the original image $I$ from the masked image $\tilde I$. However, the objective of the erasure task is to erase objects and generate a harmonious background. Because there is currently no large-scale object-level removal dataset suitable for object erasure training, we propose a new data construction strategy and build an object-level removal dataset (OLRD) based on Place2~\cite{zhou2017places}.

Specifically, given an original image $I$, we utilize a pretrained panoptic segmentation network, such as Mask2Former \cite{cheng2022masked}, to label the entire image. We then randomly select an object $o$ (e.g., ``sheep'' in Fig.~\ref{fig:data}(b)). Subsequently, the object $o$ and its mask $m$ are shifted to a region marked as background (e.g., ``grass'' or ``gravel'') based on the segmentation results, obtaining in $\tilde o$ and $\tilde m$. Finally, $\tilde o$ is blended into the original image $I$ to generate $\tilde{I}$:
\begin{equation}
    \tilde{I} = \tilde{o} + \tilde{m} * I.
\end{equation}
Our MagicEraser is obtained by fine-tuning Stable Diffusion Inpainting using $\tilde m$ and $\tilde I$ with the ground truth $I$. 
In practice, common data augmentation tricks such as scaling and rotation and color change can be applied to $\tilde o$. In this work, we do not perform them.

Additionally, to obtain the textual description of $I$, we add a prompt to guide a Vision-Language model (VLM) (e.g. LLAVA~\cite{liu2023llava}) to focus more on the background regions rather than the objects. For instance, adding a prompt like ``Describe the grass and gravel in the image'' to the VLM yields a response such as ``The grass is green and lush and the gravel is scattered throughout the scene'', placing more emphasis on the ``grass'' and ``gravel'' regions identified by panoptic segmentation. This textual description of $I$, together with $\tilde m$ and $\tilde I$, is utilized to fine-tune the T2I Stable Diffusion Inpainting model for object erasure. The optimization of the model fine-tuning using LoRA is represented in Eq.~\ref{eq:loss_v_lora}.

\section{Experiments}
\subsection{Experimental Setup}
\textbf{Implementation Details}. Our framework, MagicEraser, is built on the Stable Diffusion Inpainting model v1.5\footnote{https://github.com/runwayml/stable-diffusion}. As for the content initialization module, a traditional pretrained inpainting model Big-LaMa\footnote{https://github.com/advimman/lama} is leveraged. 
And we apply Mask2Former\footnote{https://github.com/facebookresearch/Mask2Former} to obtain the panoptic segmentation results for the semantics-aware refocus module. We use Adam optimizer with the learning rate being 1e-4 in the prompt tuning process, which takes around 50K steps.
The training data is constructed based on Place2~\cite{zhou2017places}, where the erasing masks are also generated by Mask2Former and the image captions are produced by LLaVA\footnote{https://github.com/haotian-liu/LLaVA}. All images and their corresponding masks are resized to $512\times512$ during training.



\begin{table*}[t]
	\begin{center}
 \small
		\caption{Quantitative comparison with five SOTA methods on three datasets.}
			\label{tab:quantitative}
		\setlength{\tabcolsep}{3.0mm}{
				\begin{tabular}{c|c|cccc}
					
					\toprule	Dataset&Method&PSNR$\uparrow$&SSIM$\uparrow$&LPIPS$\downarrow$&FID$\downarrow$\\
					\toprule
					\multirow{6}*{\makecell[c]{Open-\\Images}}&MAT~\cite{li2022mat}&\underline{26.994dB}&\textbf{0.949}&\textbf{0.030}&31.30 \\			
					~&Co-Mod~\cite{zhao2021large}&26.446dB&0.941&0.033&\underline{30.40} \\
                     ~&LaMa\cite{suvorov2022resolution} &21.618dB&0.936&0.055&37.10 \\
                    ~&CoordFill\cite{liu2023coordfill}&22.072dB&0.934&0.081&35.94 \\
                    ~&SD Inpainting\cite{rombach2022high}&26.096dB&0.942&0.036&31.10\\

					~&MagicEraser  &\textbf{28.123dB}&\underline{0.947}&\underline{0.032}&\textbf{30.02}\\

					\midrule
     
					\multirow{6}*{\makecell[c]{COCO}}&MAT\cite{li2022mat}&\underline{24.758}dB&\underline{0.903}&0.056&\underline{41.33} \\
     ~&Co-Mod~\cite{zhao2021large}&19.444dB&0.757&0.101&43.33 \\
     ~&LaMa\cite{suvorov2022resolution}&20.675dB&0.897&0.087&44.24 \\
					~&CoordFill \cite{liu2023coordfill}&20.966dB&0.897&\underline{0.094}&46.68 \\				
					~&SD Inpainting\cite{rombach2022high}&22.248dB&0.892&0.079&42.85\\
					~&MagicEraser&\textbf{24.766dB}&\textbf{0.908}&\textbf{0.062}&\textbf{39.55}\\

					\midrule
					
					\multirow{6}*{RealHM}&MAT\cite{li2022mat}&21.484dB&0.843&\underline{0.107}&51.73\\
                     ~&Co-Mod\cite{zhao2021large}&20.777dB&0.801&0.117&54.43 \\
                    ~&LaMa\cite{suvorov2022resolution} &19.053dB&0.825&0.150&55.70 \\
					~&CoordFill\cite{liu2023coordfill}&19.239dB&0.827&0.177&56.92 \\			
					~&SD Inpainting\cite{rombach2022high}&\underline{21.758dB}&\underline{0.846}&0.116&\textbf{45.05}\\
                    ~&MagicEraser&\textbf{23.620dB}&\textbf{0.861}&\textbf{0.101}&\underline{46.56}\\

					\bottomrule
				\end{tabular}
			}	
		\end{center}
	\end{table*}

\begin{figure*}[t]
\scriptsize
	\centering
\begin{minipage}[t]{0.1180\textwidth}	\centerline{\includegraphics[width=1.515cm]{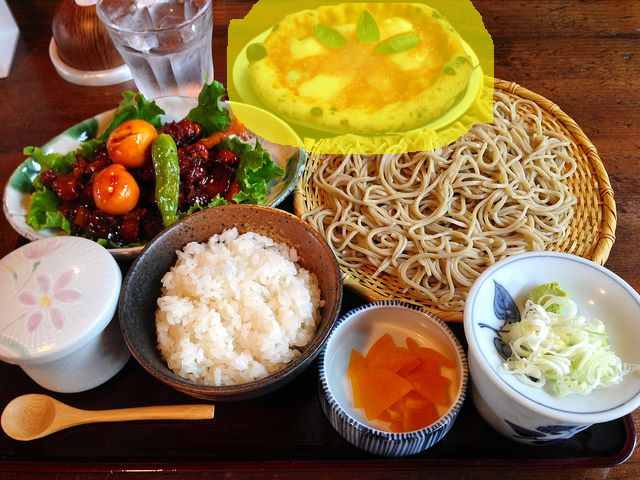}}
	\end{minipage}
\begin{minipage}[t]{0.1180\textwidth}	\centerline{\includegraphics[width=1.515cm]{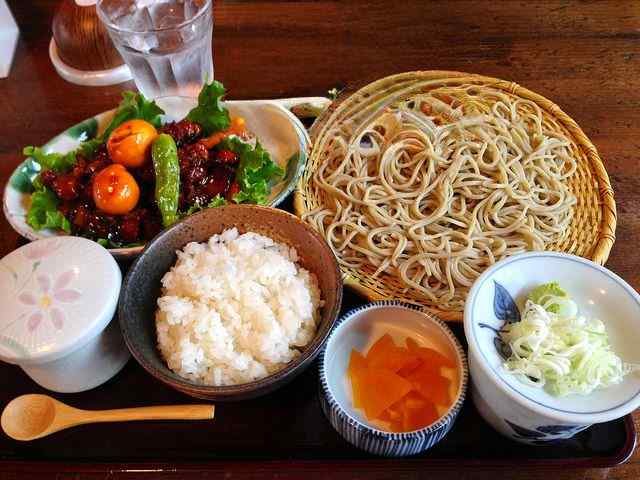}}
	\end{minipage}
\begin{minipage}[t]{0.1180\textwidth}	\centerline{\includegraphics[width=1.515cm]{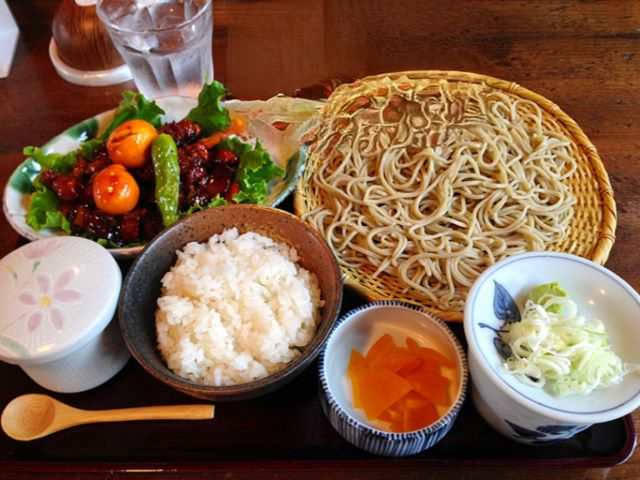}}
	\end{minipage}
\begin{minipage}[t]{0.1180\textwidth}	\centerline{\includegraphics[width=1.515cm]{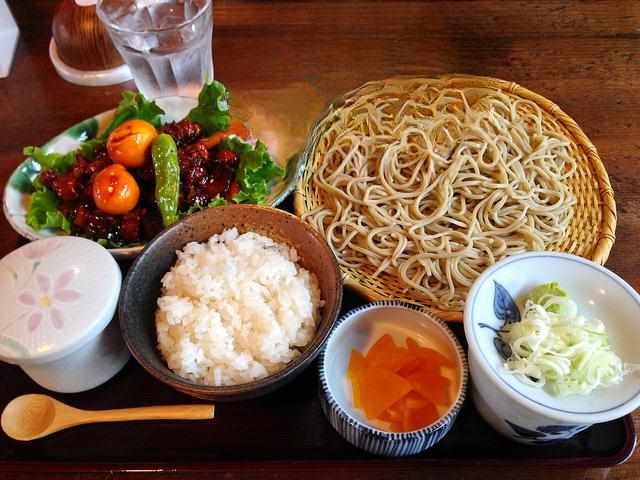}}
	\end{minipage}
\begin{minipage}[t]{0.1180\textwidth}	\centerline{\includegraphics[width=1.515cm]{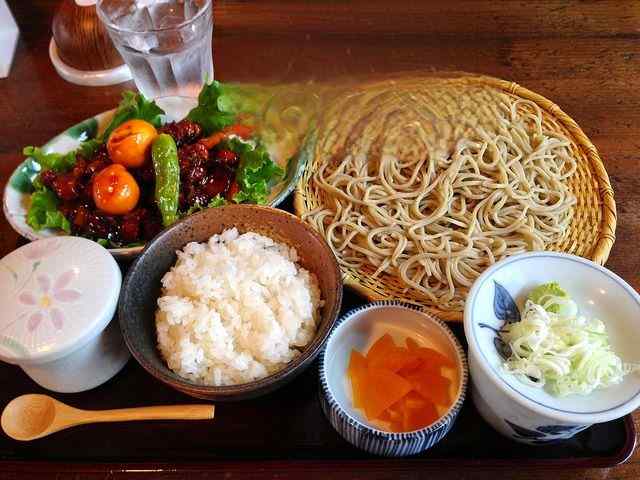}}
	\end{minipage}
\begin{minipage}[t]{0.1180\textwidth}	\centerline{\includegraphics[width=1.515cm]{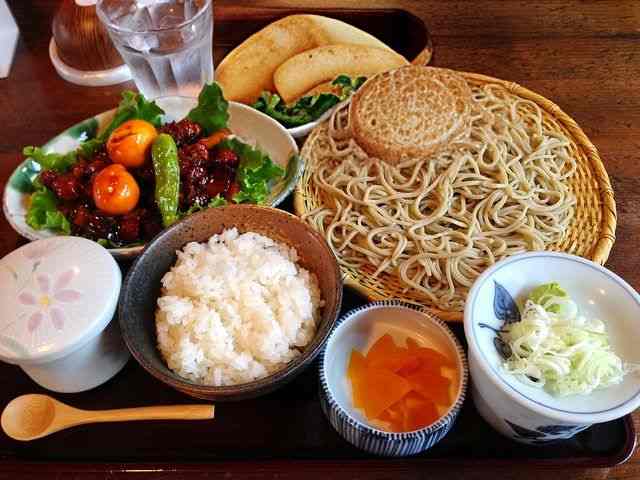}}
	\end{minipage}
\begin{minipage}[t]{0.1180\textwidth}	\centerline{\includegraphics[width=1.515cm]{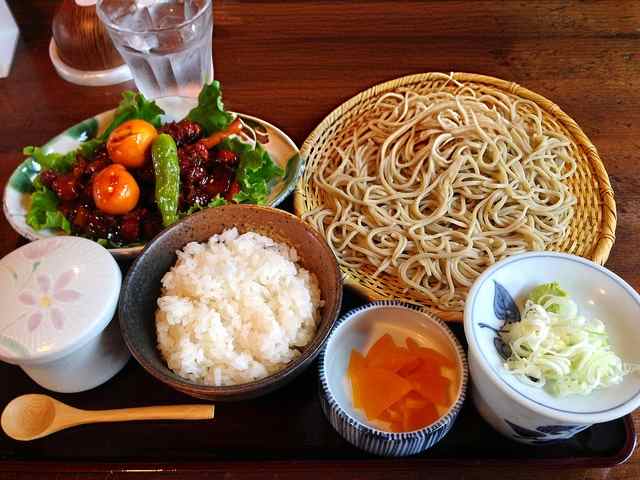}}
	\end{minipage}
\begin{minipage}[t]{0.1180\textwidth}	\centerline{\includegraphics[width=1.515cm]{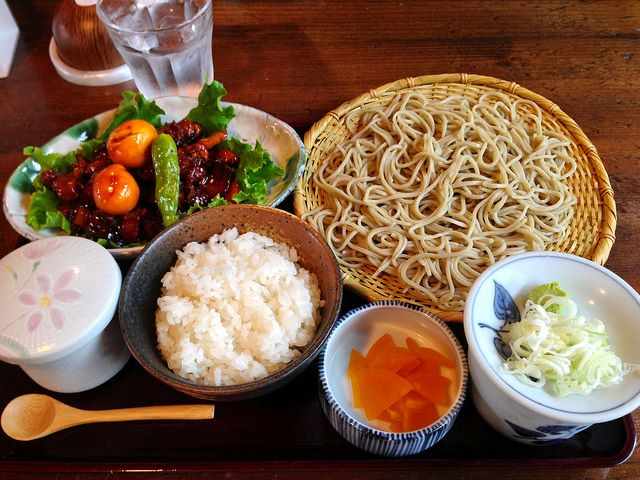}}
	\end{minipage}
\hfill
\begin{minipage}[t]{0.1180\textwidth}	\centerline{\includegraphics[width=1.515cm]{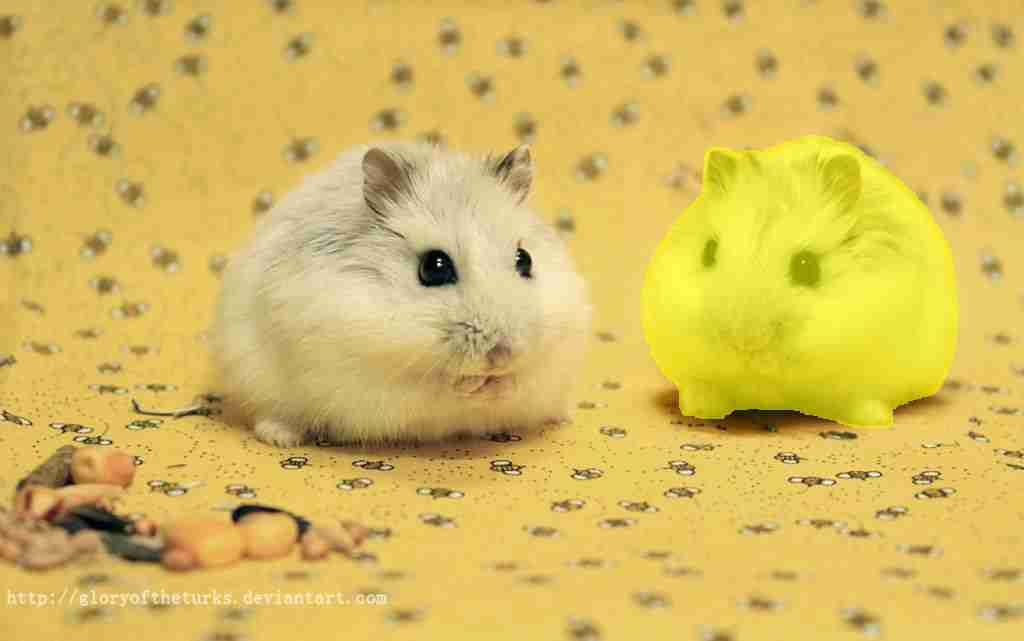}}
	\end{minipage}
\begin{minipage}[t]{0.1180\textwidth}	\centerline{\includegraphics[width=1.515cm]{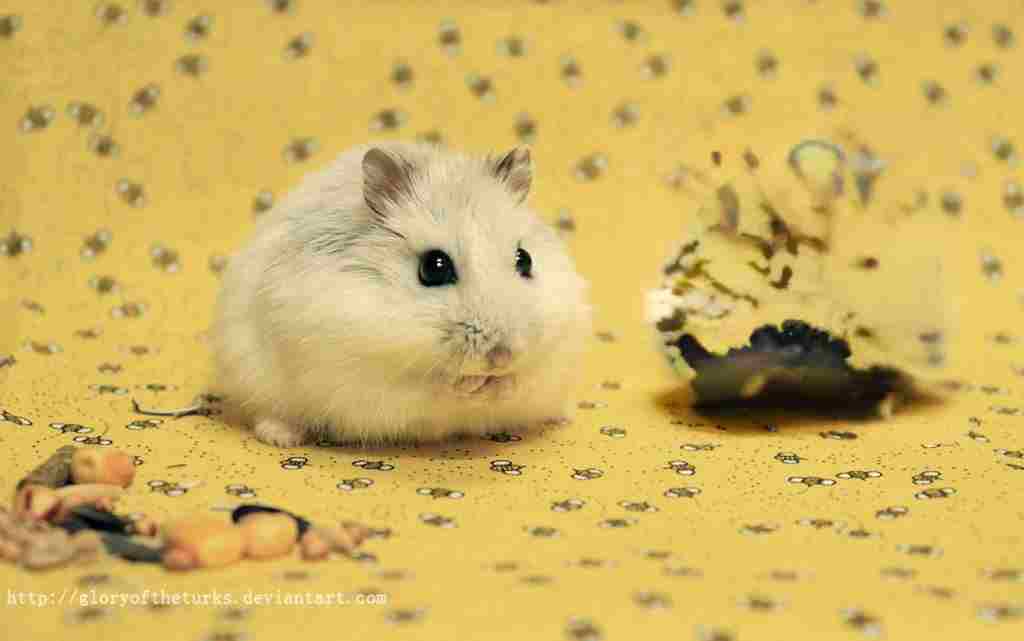}}
	\end{minipage}
\begin{minipage}[t]{0.1180\textwidth}	\centerline{\includegraphics[width=1.515cm]{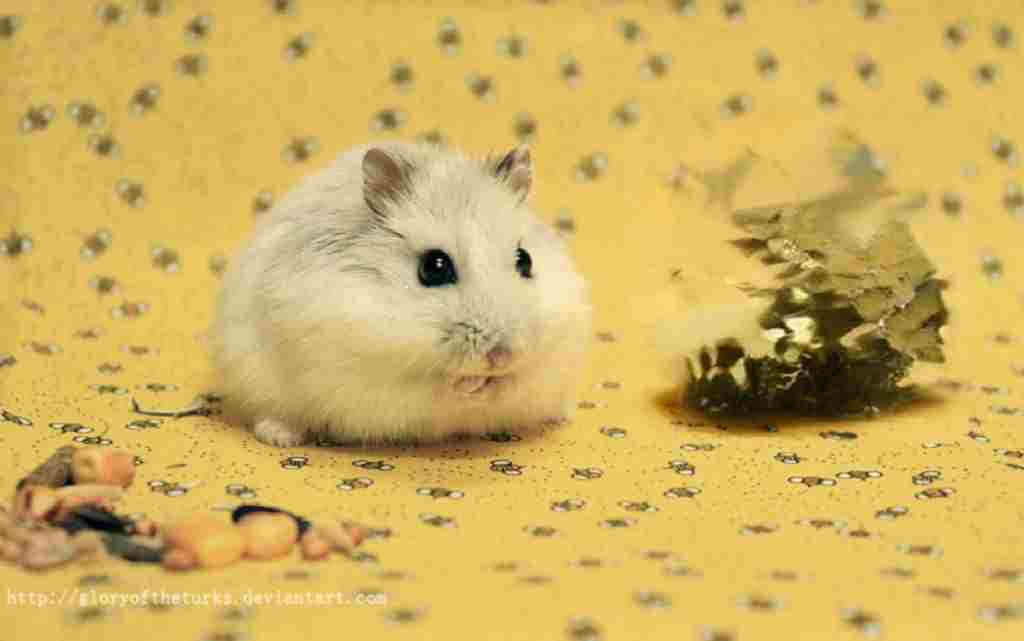}}
	\end{minipage}
\begin{minipage}[t]{0.1180\textwidth}	\centerline{\includegraphics[width=1.515cm]{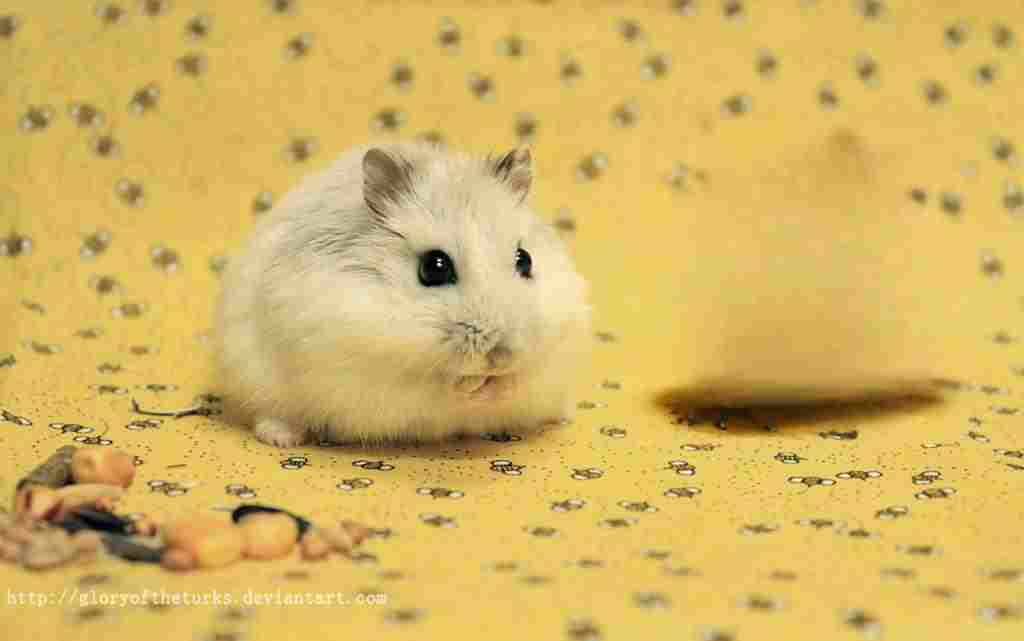}}
	\end{minipage}
\begin{minipage}[t]{0.1180\textwidth}	\centerline{\includegraphics[width=1.515cm]{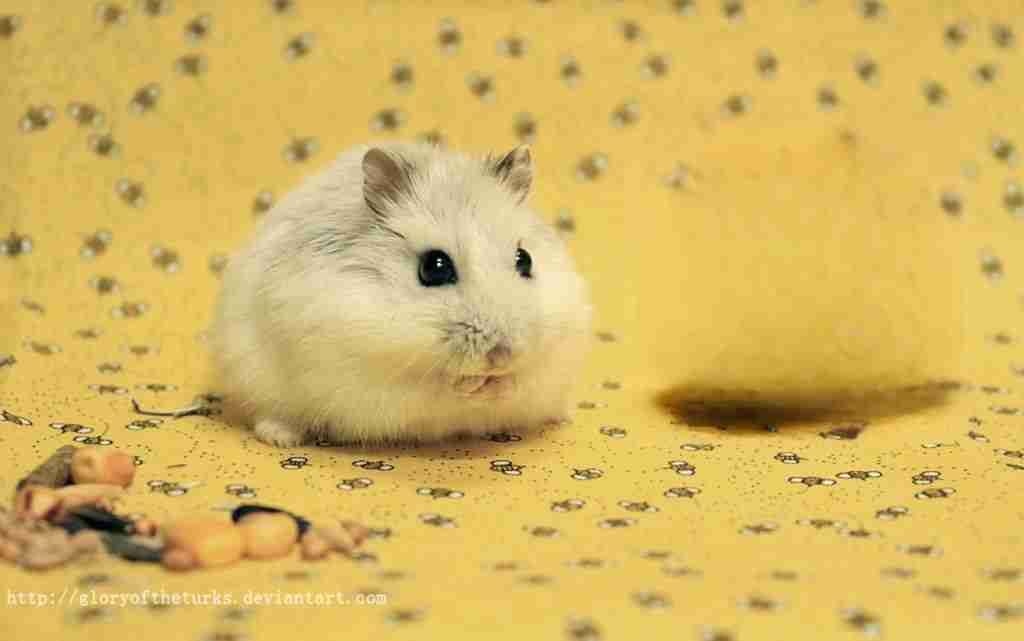}}
	\end{minipage}
\begin{minipage}[t]{0.1180\textwidth}	\centerline{\includegraphics[width=1.515cm]{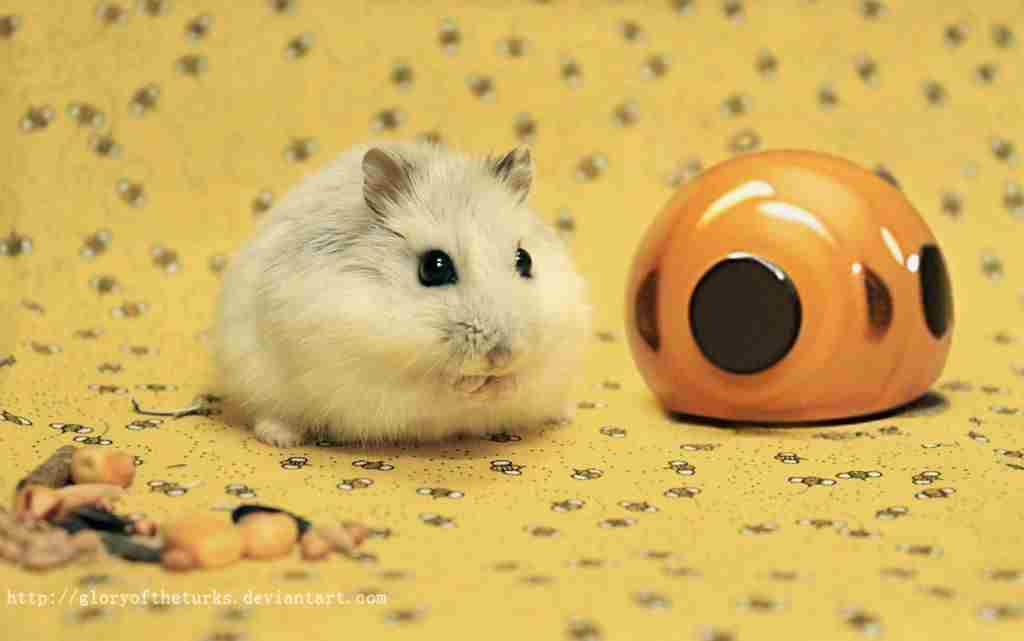}}
	\end{minipage}
\begin{minipage}[t]{0.1180\textwidth}	\centerline{\includegraphics[width=1.515cm]{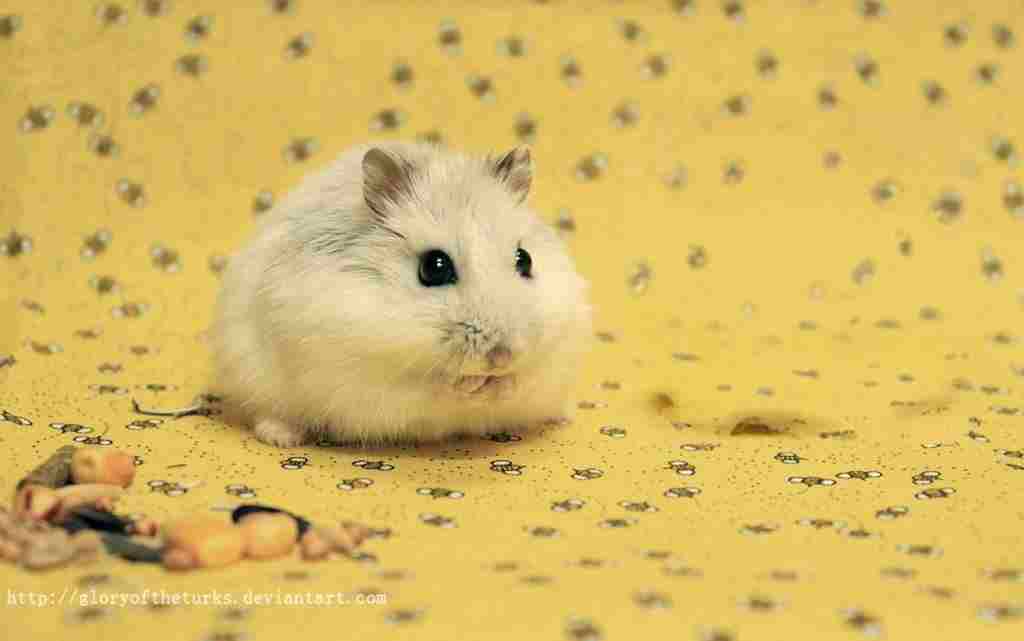}}
	\end{minipage}
\begin{minipage}[t]{0.1180\textwidth}	\centerline{\includegraphics[width=1.515cm]{fig/comp/openimages/7c33a5bb544ecf66-1_output.jpg}}
	\end{minipage}
 \hfill
\begin{minipage}[t]{0.1180\textwidth}	\centerline{\includegraphics[width=1.515cm]{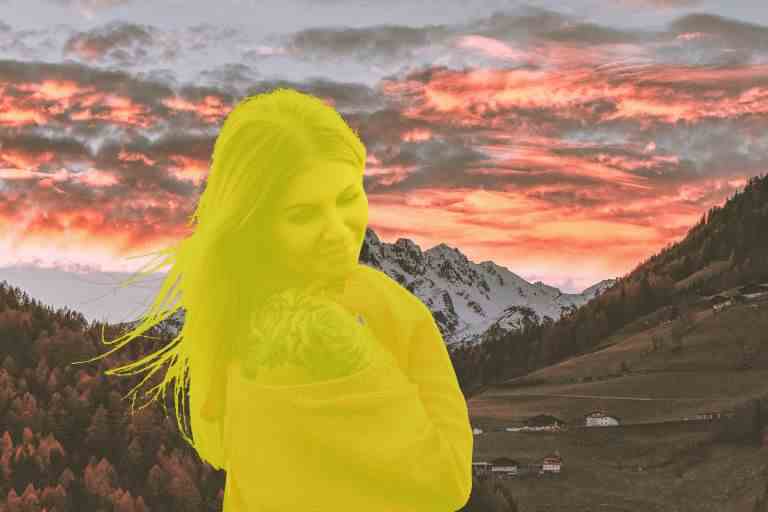}}
	\end{minipage}
\begin{minipage}[t]{0.1180\textwidth}	\centerline{\includegraphics[width=1.515cm]{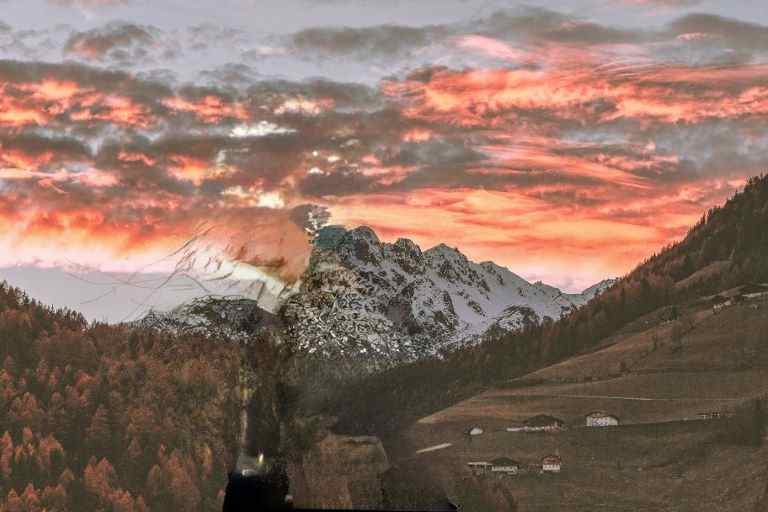}}
	\end{minipage}
\begin{minipage}[t]{0.1180\textwidth}	\centerline{\includegraphics[width=1.515cm]{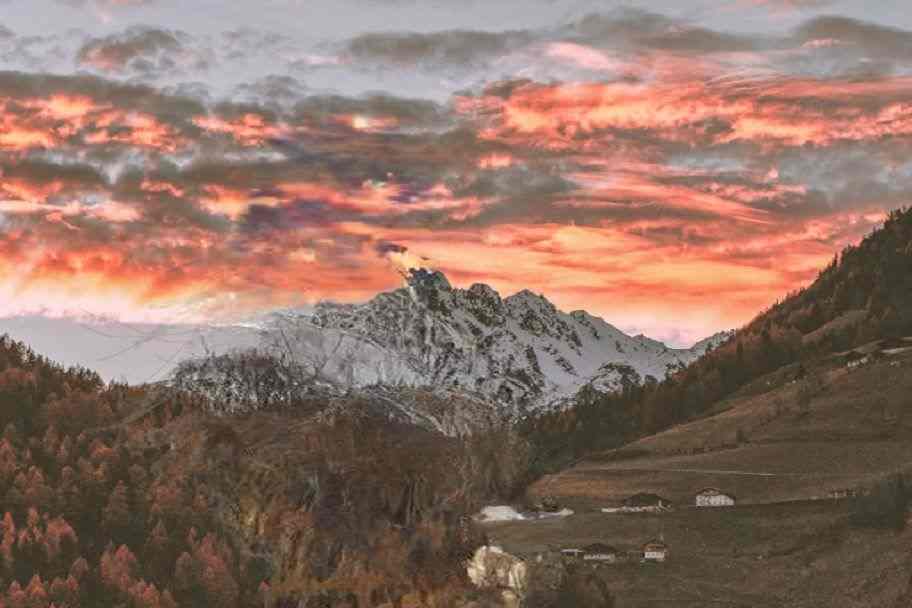}}
	\end{minipage}
\begin{minipage}[t]{0.1180\textwidth}	\centerline{\includegraphics[width=1.515cm]{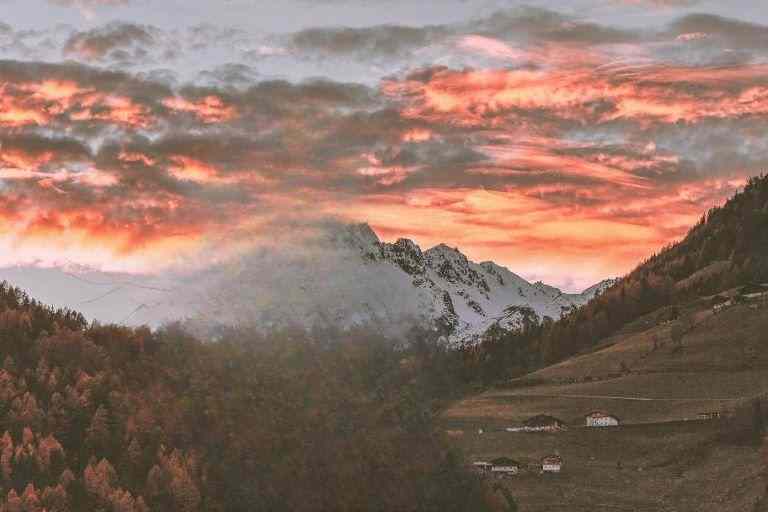}}
\end{minipage}
\begin{minipage}[t]{0.1180\textwidth}	\centerline{\includegraphics[width=1.515cm]{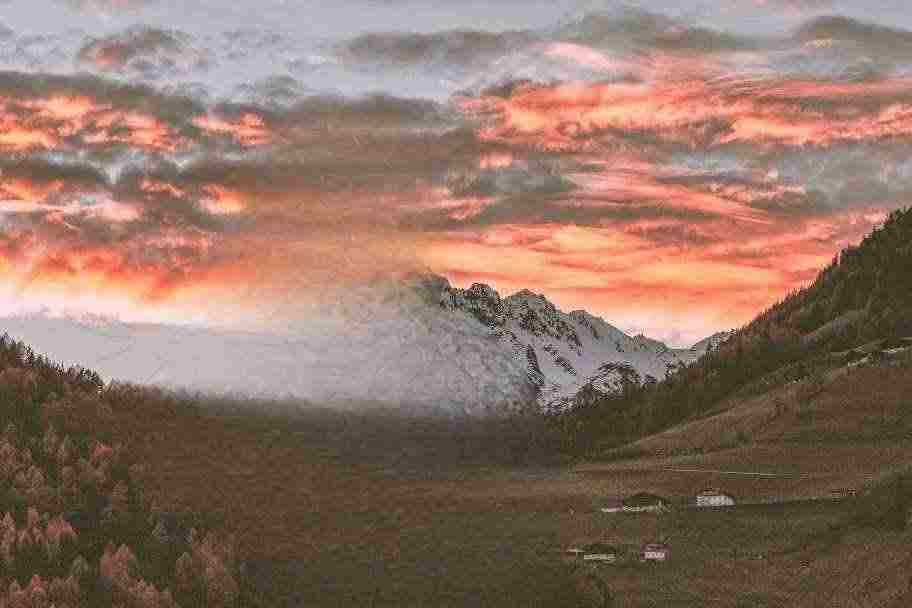}}
	\end{minipage}
\begin{minipage}[t]{0.1180\textwidth}	\centerline{\includegraphics[width=1.515cm]{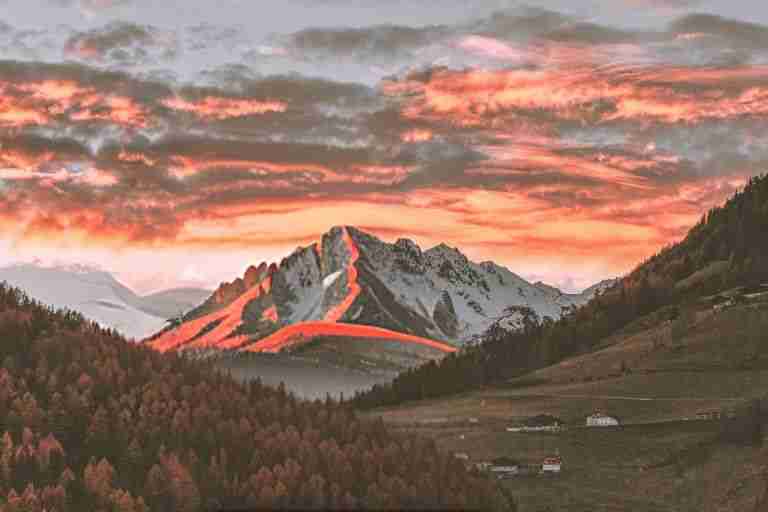}}
	\end{minipage}
\begin{minipage}[t]{0.1180\textwidth}	\centerline{\includegraphics[width=1.515cm]{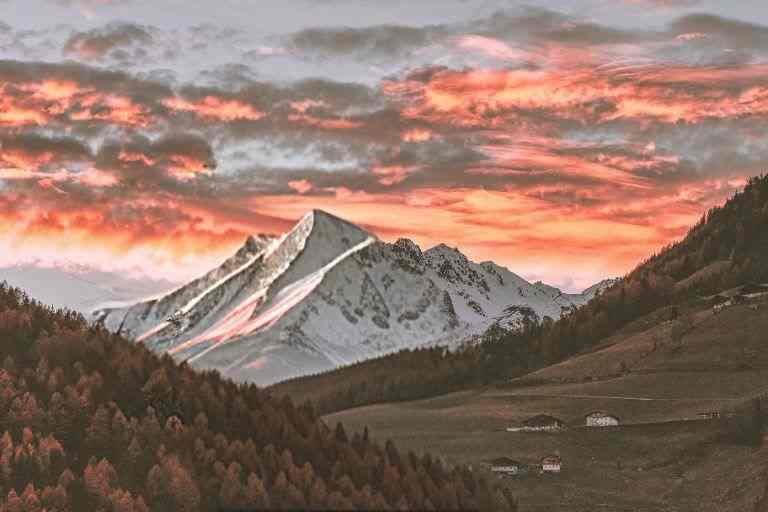}}
	\end{minipage}
\begin{minipage}[t]{0.1180\textwidth}	\centerline{\includegraphics[width=1.515cm]{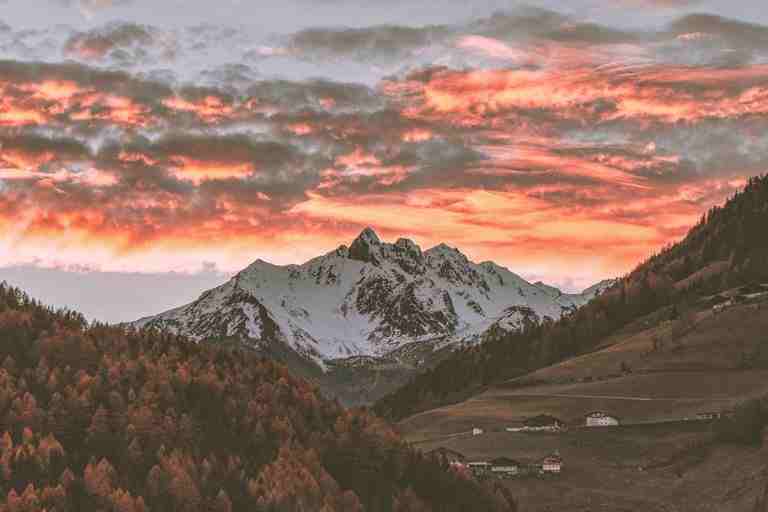}}
	\end{minipage}
 \hfill
\begin{minipage}[t]{0.1180\textwidth}	\centerline{\includegraphics[width=1.515cm]{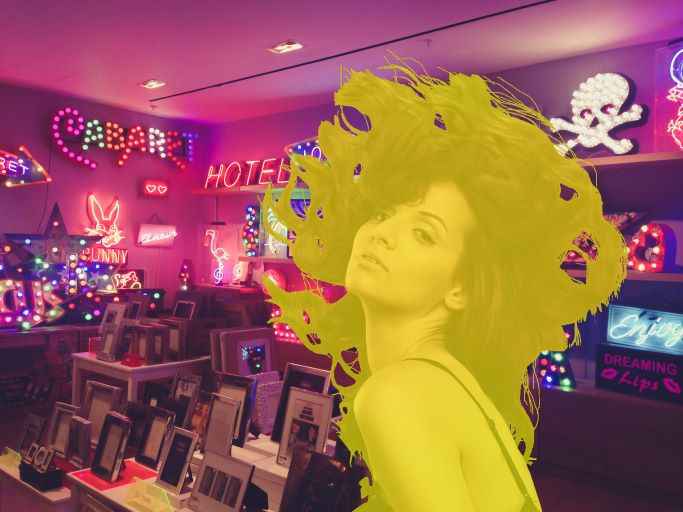}}
	\end{minipage}
\begin{minipage}[t]{0.1180\textwidth}	\centerline{\includegraphics[width=1.515cm]{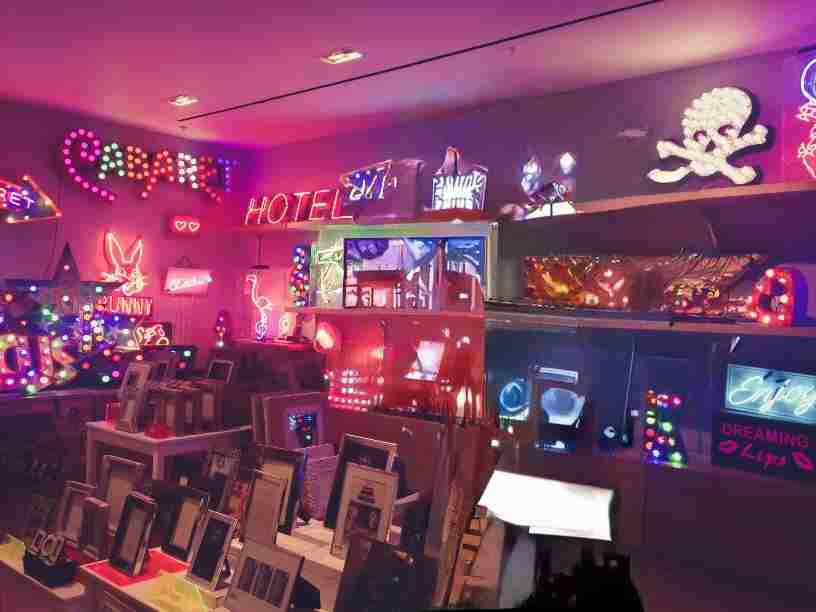}}
	\end{minipage}
\begin{minipage}[t]{0.1180\textwidth}	\centerline{\includegraphics[width=1.515cm]{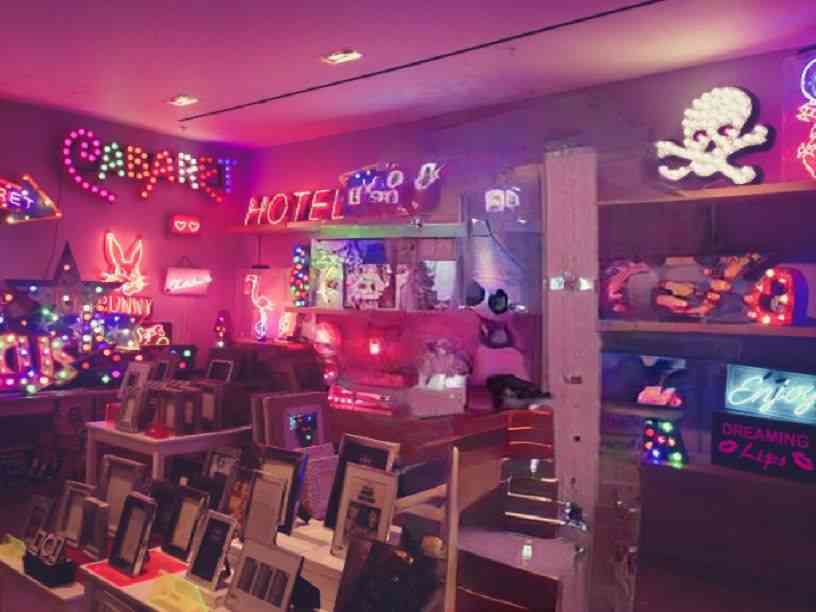}}
	\end{minipage}
\begin{minipage}[t]{0.1180\textwidth}	\centerline{\includegraphics[width=1.515cm]{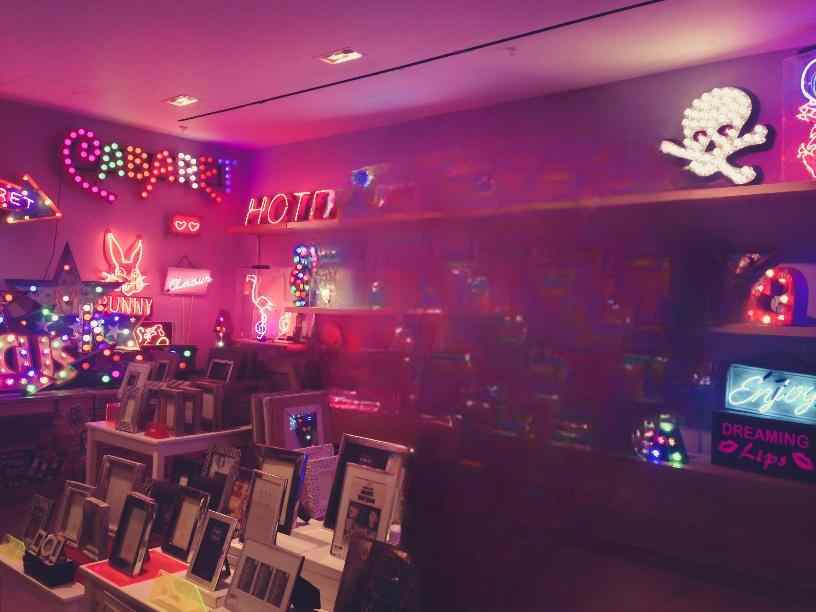}}
\end{minipage}
\begin{minipage}[t]{0.1180\textwidth}	\centerline{\includegraphics[width=1.515cm]{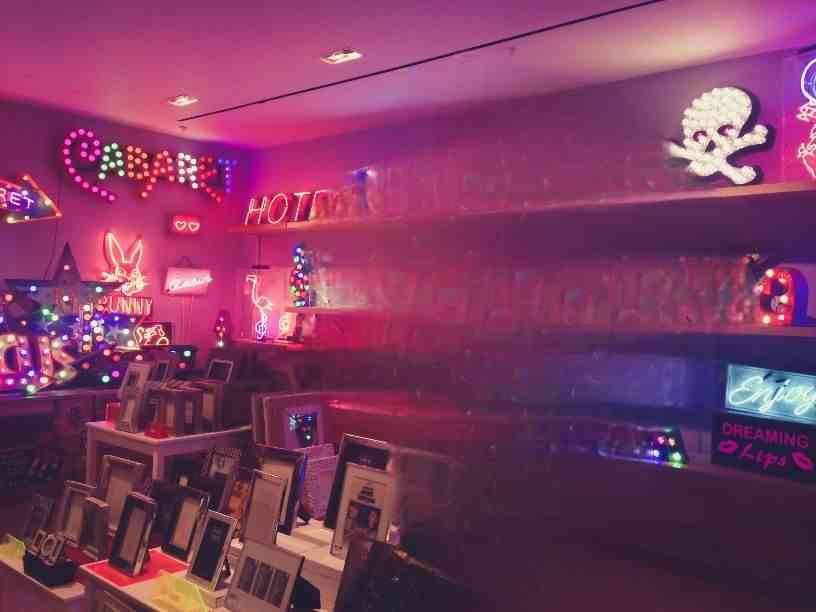}}
	\end{minipage}
\begin{minipage}[t]{0.1180\textwidth}	\centerline{\includegraphics[width=1.515cm]{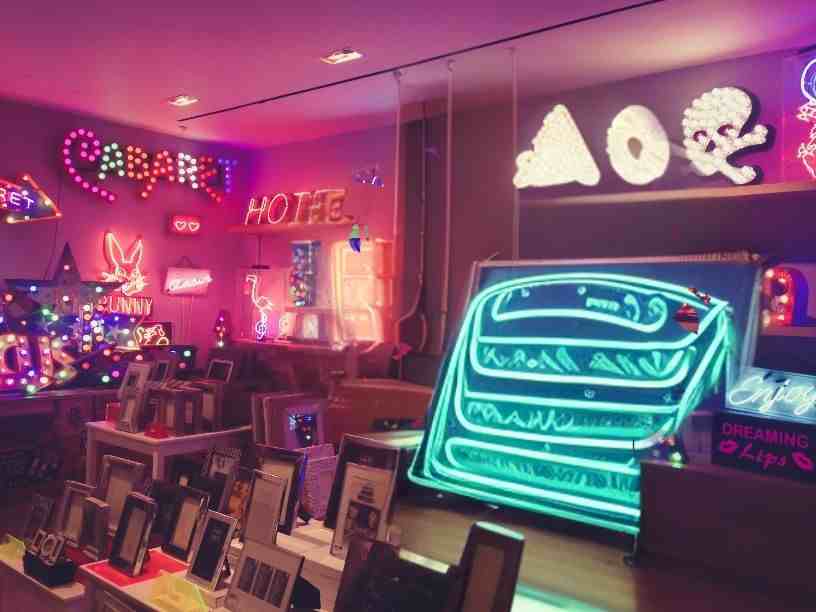}}
	\end{minipage}
\begin{minipage}[t]{0.1180\textwidth}	\centerline{\includegraphics[width=1.515cm]{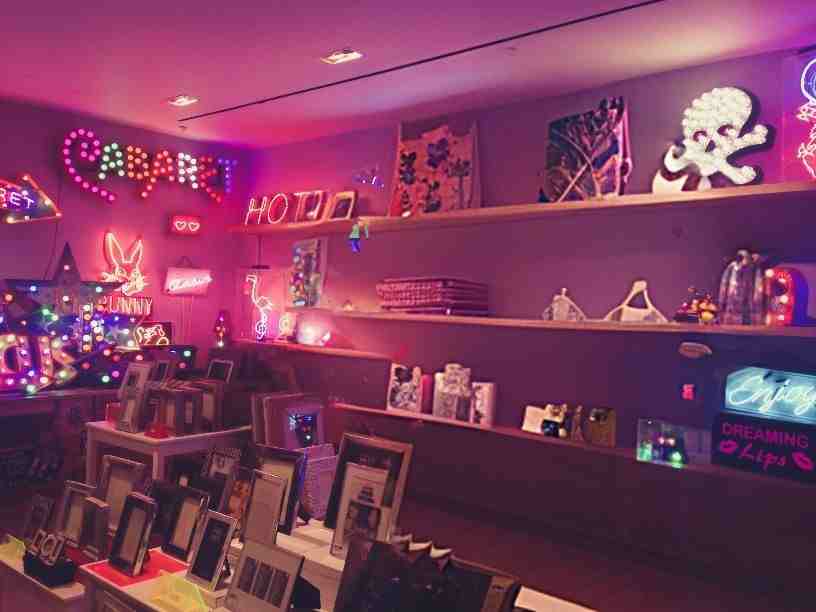}}
	\end{minipage}
\begin{minipage}[t]{0.1180\textwidth}	\centerline{\includegraphics[width=1.515cm]{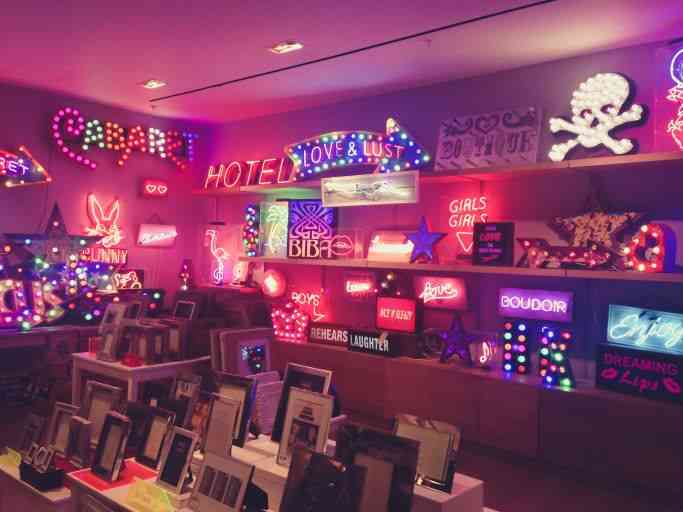}}
	\end{minipage}
 \hfill
\begin{minipage}[t]{0.1180\textwidth}	\centerline{\includegraphics[width=1.515cm]{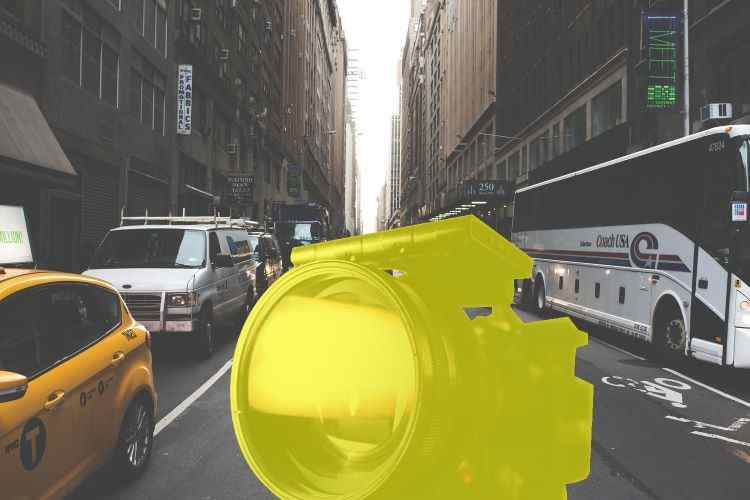}}
 \centerline{Input+Mask}
	\end{minipage}
\begin{minipage}[t]{0.1180\textwidth}	\centerline{\includegraphics[width=1.515cm]{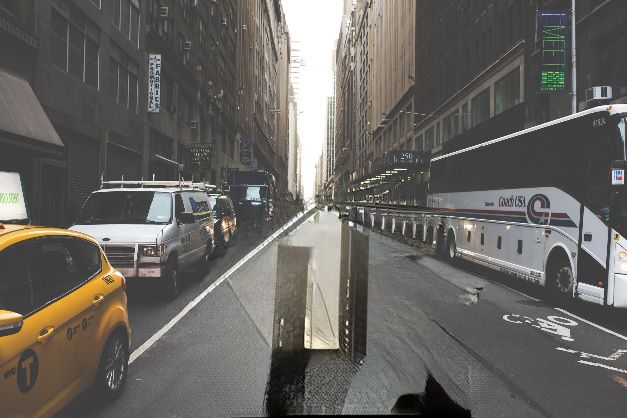}}
	\centerline{MAT}
	\end{minipage}
\begin{minipage}[t]{0.1180\textwidth}	\centerline{\includegraphics[width=1.515cm]{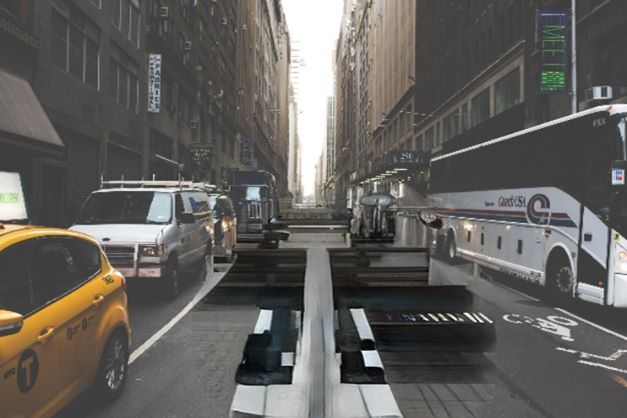}}
 \centerline{Co-Mod}
	\end{minipage}
\begin{minipage}[t]{0.1180\textwidth}	\centerline{\includegraphics[width=1.515cm]{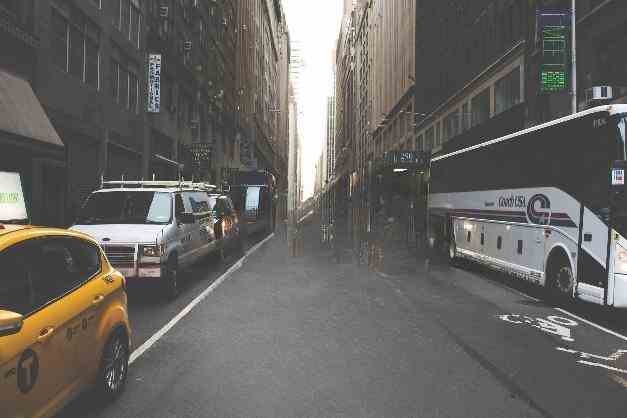}}
    \centerline{LaMa}
	\end{minipage}
\begin{minipage}[t]{0.1180\textwidth}	\centerline{\includegraphics[width=1.515cm]{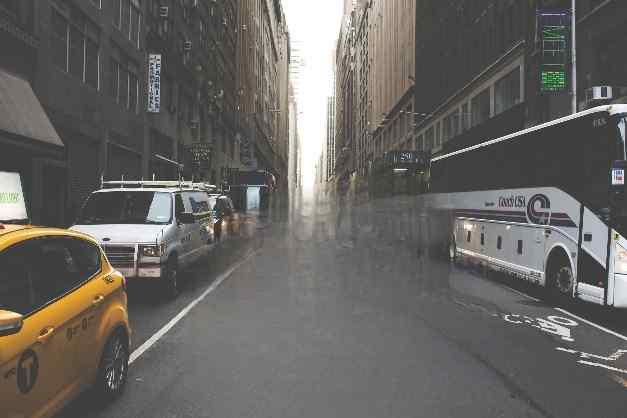}}
 \centerline{CoordFill}
	\end{minipage}
\begin{minipage}[t]{0.1180\textwidth}	\centerline{\includegraphics[width=1.515cm]{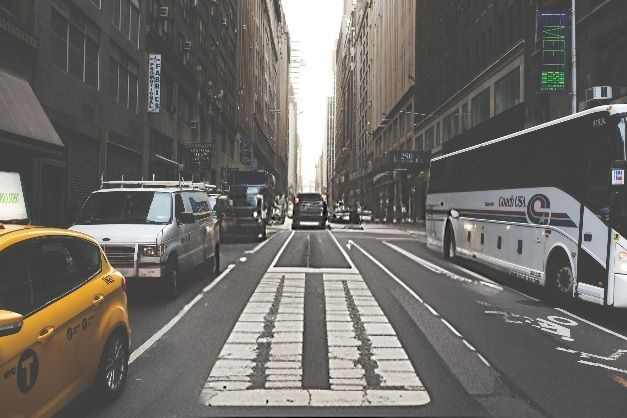}}
\centerline{SD} 
\centerline{Inpainting}
	\end{minipage}
\begin{minipage}[t]{0.1180\textwidth}	\centerline{\includegraphics[width=1.515cm]{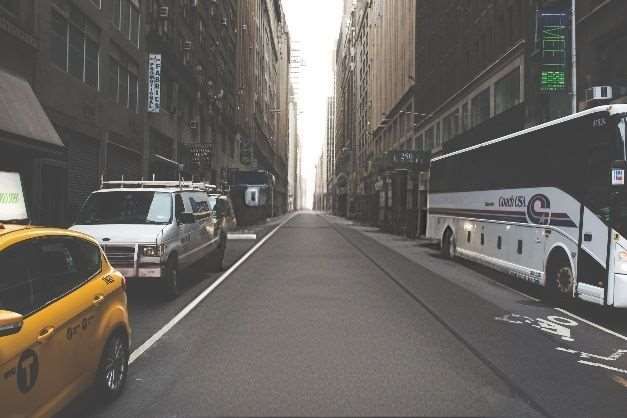}}
\centerline{MagicEraser}
	\end{minipage}
\begin{minipage}[t]{0.1180\textwidth}	\centerline{\includegraphics[width=1.515cm]{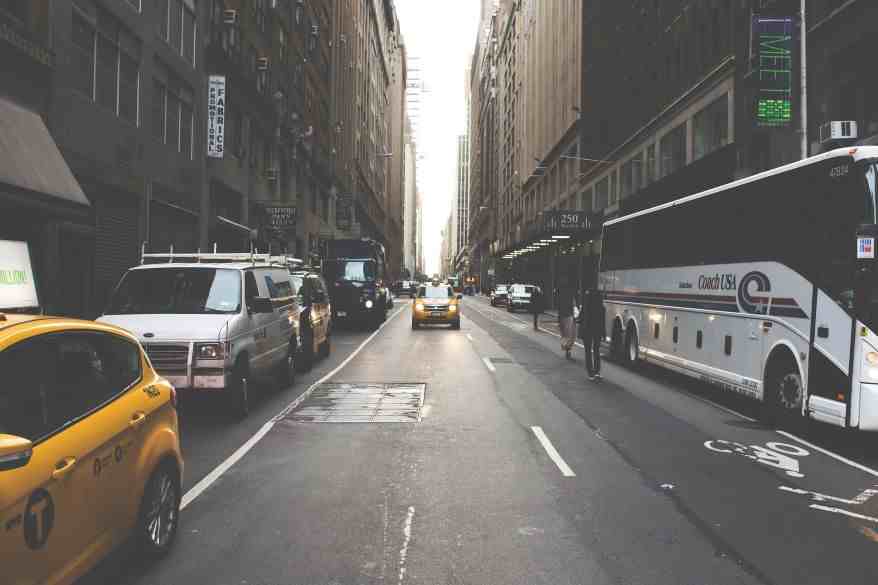}}
\centerline{Reference}
	\end{minipage}
\caption{Visual comparison with five SOTA algorithms.}
	\label{fig: main-result} 
\end{figure*}

\subsubsection{Evaluation Datasets and Metrics.} We assess the performance of our MagicEraser on three different datasets: OpenImages~\cite{OpenImages}, COCO~\cite{lin2014microsoft} and RealHM~\cite{jiang2021ssh}. From OpenImages and COCO, we respectively sample 200 representative examples with side resolution higher than 512 and then construct the erasing pairs with the strategy detailed in Section~\ref{sec:data}. As for RealHM, it is collected for self-supervised image harmonization, containing 215 high-quality examples with side resolution higher than 4000. Specifically, every example of RealHM already contains an original image $I$, an object mask $m$ and a blended result $\tilde I$ which can directly be applied to assess the performance of the erasure task by using $\tilde I$ and $m$ to recover $I$. During testing, every image is transformed to the size of $512\times512$ by linearly mapping its long side to 512 and then padding the short side with values 0. By comparing the erasure results with their corresponding reference images (i.e., the original images), we report PSNR~\cite{wang2004image}, SSIM~\cite{wang2004image}, LPIPS~\cite{zhang2018unreasonable}, and FID~\cite{heusel2017gans} as the quantitative evaluation metrics.



\subsection{Comparison with State-of-the-Arts} 

To evaluate the effectiveness of MagicEraser, we conduct a comprehensive comparison with state-of-the-art (SOTA) methods in the field of image inpainting, including four traditional GAN-based approaches (MAT~\cite{li2022mat}, Co-Mod~\cite{zhao2021large}, LaMa~\cite{suvorov2022resolution} and CoordFill~\cite{liu2023coordfill}) and a diffusion model-based method SD Inpainting~\cite{rombach2022high}. We utilize LLaVA~\cite{liu2023llava} to craft detailed textual prompts for SD Inpainting.
\Cref{tab:quantitative} lists the quantitative results of the compared methods across four metrics. It shows that MagicEraser outperforms others by a large margin in terms of PSNR and obtains competitive performance in SSIM, indicating its superior effectiveness in erasing objects and recovering backgrounds. Furthermore, MagicEraser excels in LPIPS and FID, demonstrating its capability to maintain visual reality and aesthetic quality while effectively removing objects from images. This conclusion can also be validated in the visual comparison of Fig.~\ref{fig: main-result}. Through this comparison, we observe distinct limitations in the performance of other methods. MAT and Co-Mod fail to completely erase the masked objects, resulting in ghost remnants of the objects or the introduction of artifacts (see all the rows). LaMa and CoordFill tend to induce severe blurriness within the masked regions, particularly in scenes with complex textures (see all the rows).
While SD Inpainting shows an improvement over the aforementioned GAN-based approaches by avoiding such artifacts and blurriness, it struggles with instability and sometimes generates unwanted elements unrelated to the original backgrounds (see the second row). Compared with them, our MagicEraser can stably generate highly realistic content harmonious with the surrounding context and obtain the most visually pleasing results.

\begin{table*}[t]
\small
\centering
\caption{Quantitative comparison with two commercial products on RealHM.}
\setlength{\tabcolsep}{2.25mm}{
\begin{tabular}{c|c c c c c c c c}
\toprule
Model &PSNR$\uparrow$&SSIM$\uparrow$&LPIPS$\downarrow$&FID$\downarrow$\\
\toprule
Adobe PhotoShop's Generative Fill &22.913dB&0.851 &0.113 & 49.07 \\
Google Photos Eraser &20.310dB&0.822 &0.173 &53.55 \\
\toprule
MagicEraser&\textbf{23.620dB}&\textbf{0.861}&\textbf{0.101}&\textbf{46.56} \\
\bottomrule
\end{tabular}}
\label{tab:comp_product}
\end{table*}

Moreover, we also compare MagicEraser with two commercial products, Adobe Photoshop’s Generative Fill\footnote{https://www.adobe.com/products/firefly.html, May 11, 2024} and Google Photos  Eraser\footnote{Google Pixel8 Build Number AP1A.240305.019.A1}. The quantitative results on the RealHM dataset are shown in Table~\ref{tab:comp_product}, which  demonstrate that our method achieves better performance.

\subsection{Ablation Study}
We perform a comprehensive ablation study to assess the impact of each component in MagicEraser on RealHM with $512\times512$ images. The quantitative results are listed in~\Cref{tab:abla_rebuttal}, where the baseline is Stable Diffusion Inpainting. 

Comparing (\romannumeral1) and (\romannumeral2), we see that our dataset OLRD achieves better performance than traditional random mask training. This is because during model training, the random masking scheme in the traditional inpainting task often leads to recovering the missing regions no matter they are background or objects, which
is not suitable for object erasure. Comparing (\romannumeral2) and (\romannumeral3), we see an increase of PSNR around 0.7dB when content initialization is employed. Because the content initialization utilizes the traditional pretrained GAN-based method to initialize the latent of Stable Diffusion, the model without it generates the images from random noise, which easily leads to unwanted artifacts. Comparing (\romannumeral3), (\romannumeral4), and (\romannumeral5), both Prompt Tuning and Semantics-Aware Attention Refocus further improve the model's performance.
Moreover, based on the ablation results (\romannumeral4) and (\romannumeral5) in Table~\ref{tab:abla_rebuttal}, Prompt Tuning is more important than Semantics-Aware Attention Refocus. While both modules help the diffusion model utilize background information to fill masked regions, they work differently. Prompt Tuning globally encodes the semantic clues through the learnable text embedding and LoRA to guide content generation aligned with the overall background concept. Semantics-Aware Attention Refocus locally modulates self-attentions to generate content spatially consistent with the background.

These results demonstrate that the three proposed components are vital to our MagicEraser framework and all have obvious contributions.

\begin{table}[t]
\centering
\caption{Ablation study on the RealHM dataset with 512 × 512 images.}
\resizebox{\linewidth}{!}{
\begin{tabular}{l|c c c c c c c c}
\toprule
Model &PSNR$\uparrow$&SSIM$\uparrow$&LPIPS$\downarrow$&FID$\downarrow$\\
\toprule
\romannumeral1. Baseline + Traditional Random Mask Traning &21.331dB&0.815&0.134&52.10 \\
\romannumeral2. Baseline + OLRD &22.130dB&0.834&0.119& 50.73 \\
\romannumeral3. Baseline + OLRD + Content Initialization &22.891dB&0.840 &0.109 & 48.91 \\
\romannumeral4. Baseline + OLRD + Content Initialization + Semantics-Aware Attention Refocus &23.277dB&0.844&0.110&48.93 \\
\romannumeral5. Baseline + OLRD + Content Initialization + Prompt Tuning &\underline{23.311dB}&\underline{0.858}&\underline{0.104}&\underline{47.94}\\
\toprule
\romannumeral6. Baseline + OLRD + Content Initialization + Prompt Tuning + Semantics-Aware Attention Refocus\\(MagicEraser)&\textbf{23.620dB}&\textbf{0.861}&\textbf{0.101}&\textbf{46.56} \\
\bottomrule
\end{tabular}}
\label{tab:abla_rebuttal}
\end{table}

\section{Limitation and Conclusion}
Although notable advantages are demonstrated by our proposed framework, there are still some limitations. Following Stable Diffusion v1.5, MagicEraser works with $512\times512$ images, where the original high-frequency details of high-resolution images (e.g., 2k, 4k and 8k) may not be preserved. On the other hand, the semantics-aware attention refocus module is sensitive to the results of the pretrained segmentation model. For example, if the background region is not properly segmented, the generated content may appear discordant.  

We have proposed a diffusion model-based framework MagicEraser especially suitable for the object erasure task which is recently in increasing demand. It utilizes a traditional inpainting algorithm to roughly initialize the content, and leverages the significant generation capacity of Stable Diffusion by fine-tuning the model with a new dataset OLRD. To further control the generation, we develop a universal prompt tuning module and a semantics-aware attention refocus module. The experiments show that MagicEraser performs best on several datasets compared with several state-of-the-art methods.


%
%
\bibliographystyle{splncs04}
\bibliography{main}
\end{document}